\PassOptionsToPackage{dvipsnames}{xcolor}
\documentclass[sigconf,screen]{acmart}

\setcopyright{none}
\renewcommand\footnotetextcopyrightpermission[1]{}
\AtBeginDocument{%
  }

\setcopyright{acmlicensed}
\copyrightyear{2018}
\acmYear{2018}
\acmDOI{XXXXXXX.XXXXXXX}
\acmConference[ACM MM]{Make sure to enter the correct
  conference title from your rights confirmation email}{October 27--31,
  2025}{Dublin, Ireland}
\acmISBN{978-1-4503-XXXX-X/2018/06}

\usepackage{subfig}
\usepackage{tikz}
\usetikzlibrary{spy,backgrounds}

\usepackage[skip=2pt]{caption}
\usepackage{setspace}
\usepackage{enumitem}

\setlist{nosep} 
\newcommand{\setvspace}[2]{%
  #1 = #2
  \advance #1 by -1\parskip}

\makeatletter 
\def\thm@space@setup{%
  \thm@preskip=3pt
  \thm@postskip=\thm@preskip 
}
\makeatother

\setlist[itemize]{noitemsep, topsep=0pt}

\makeatletter
\g@addto@macro\normalsize{%
  \setlength\abovedisplayskip{1pt}
  \setlength\belowdisplayskip{1pt}
  \setlength\abovedisplayshortskip{1pt}
  \setlength\belowdisplayshortskip{1pt}
}
\makeatother

\makeatletter 
\renewenvironment{proof}[1][\proofname]{\par
  \vspace{1pt}
  \pushQED{\qed}%
  \normalfont
  \topsep0pt \partopsep0pt 
  \trivlist
  \item[\hskip\labelsep
        \itshape
    #1\@addpunct{.}]\ignorespaces
}{%
  \popQED\endtrivlist\@endpefalse
  \addvspace{3pt plus 3pt} 
}
\makeatother 

\NewDocumentCommand{\var}{O{s} m O{}}{%
  \ensuremath{#1_{#2}^{#3}}
}
\usepackage{siunitx}
\usepackage{colortbl}
\usepackage{xspace}

\newcommand{\commentout}[1]{}

\definecolor{light-gray}{gray}{0.80}

\newcommand\fref{Fig.~\ref}
\newcommand\tref{Tbl.~\ref}
\newcommand\sref{\S~\ref}

\usepackage{amsthm}

\usepackage{amsthm}

\newtheorem{mytheorem}{Theorem}
\newtheorem{assumption}[mytheorem]{Assumption}
\newtheorem{lemma}[mytheorem]{Lemma}
\newtheorem{remk}[mytheorem]{Remark}

\newcommand{\name}{FlexGaussian\xspace}
\newcommand{\gaussian}{3D-GS\xspace}
\newcommand{\pruning}{Attribute-Discriminative Pruning\xspace}

\newcommand{\adp}{ADP\xspace}
\newcommand{\mpq}{MPQ\xspace}
\newcommand{\foa}{FOA\xspace}
\usepackage{amsmath}
\usepackage{bbm}
\usepackage{multirow}

\newcommand{\zoomin}[9]{ %
\begin{tikzpicture}[spy using outlines={rectangle,#9,magnification=#8,size=#6}]   
	\node[anchor=south west,inner sep=0]  {\includegraphics[width=#7]{#1}};
	\spy on (#2, #3) in node at (#4,#5);
\end{tikzpicture}
}

\usepackage{threeparttable}

\ifx\figurename\undefined \def\figurename{Figure}\fi
\renewcommand{\figurename}{Fig.}

\newcommand{\bestcell}[1]{\colorbox{ForestGreen!50}{#1}}
\newcommand{\oom}[1]{\colorbox{VioletRed!90}{OOM}}

\newcommand{\best}[1]{\colorbox{Green!20}{#1}}
\newcommand{\second}[1]{\colorbox{Cyan!30}{#1}}
\newcommand{\third}[1]{\colorbox{Orange!30}{#1}}

\newcommand{\fixme}[1]{{{#1}}}

\begin{document}

\title{\name: Flexible and Cost-Effective Training-Free Compression for 3D Gaussian Splatting}


\settopmatter{authorsperrow=5}

\author{Boyuan Tian}
\email{boyuant2@illinois.edu}
\affiliation{
    \institution{UIUC}
    \country{}
}

\author{Qizhe Gao}
\email{qizheg2@illinois.edu}
\affiliation{
    \institution{UIUC}
    \country{}
}

\author{Siran Xianyu}
\email{sxianyu2@illinois.edu}
\affiliation{
    \institution{UIUC}
    \country{}
}

\author{Xiaotong Cui}
\email{xcui15@illinois.edu}
\affiliation{
    \institution{UIUC}
    \country{}
}

\author{Minjia Zhang}
\email{minjiaz@illinois.edu}
\affiliation{
    \institution{UIUC}
    \country{}
}

\begin{abstract}
3D Gaussian splatting has become a prominent technique for representing and rendering complex 3D scenes, due to its high fidelity and speed advantages. However, the growing demand for large-scale models calls for effective compression to reduce memory and computation costs, especially on mobile and edge devices with limited resources. Existing compression methods effectively reduce 3D Gaussian parameters but often require extensive retraining or fine-tuning, lacking flexibility under varying compression constraints.

In this paper, we introduce \name, a flexible and cost-effective method that combines mixed-precision quantization with attribute-discriminative pruning for training-free 3D Gaussian compression. \name eliminates the need for retraining and adapts easily to diverse compression targets. Evaluation results show that \name achieves up to 
96.4\%
compression while maintaining high rendering quality ($<$1 dB drop in PSNR), and is deployable on mobile devices. \name delivers high compression ratios within seconds, being 1.7-2.1$\times$ faster than state-of-the-art training-free methods and 10-100$\times$ faster than training-involved approaches. The code is being prepared and will be released soon at: \href{https://github.com/Supercomputing-System-AI-Lab/FlexGaussian}{https://github.com/Supercomputing-System-AI-Lab/FlexGaussian}
\end{abstract}






\maketitle

\section{Introduction}
\label{sec:intro}

3D Gaussian splatting (\gaussian) has emerged as a powerful technique for novel view synthesis (NVS)~\cite{mildenhall2021nerf, kerbl20233d}, which generates new views of a 3D scene by interpolating from a small set of images with known camera parameters. 
By leveraging point-based representations (e.g., Gaussian primitives) to model scene geometry and appearance, \gaussian delivers enhanced realism, detailed lighting information, and significantly faster rendering speeds than alternative Neural Radiance Fields (NeRF) methods~\cite{mildenhall2021nerf, mip-nerf360, mueller2022instant}, making it a promising candidate for enhancing applications in mobile, gaming, and AR/VR devices. However, the heavy storage requirements for Gaussians often hinder the deployment of 3D-GS models on hardware platforms with limited resources, particularly when representing large-scale, high-resolution scenes that naturally involve more Gaussians. For intuitive illustration, the \textsc{Garden} scene from the Mip-NeRF360 dataset, covering an area of less than 100$m^2$ and trained at 1.6$K$ image resolution, requires over 5.8 million explicit Gaussian primitives, demanding gigabyte level of storage~\cite{kerbl20233d}.

\begin{figure}[t]
    \centering
    \includegraphics[width=\columnwidth]{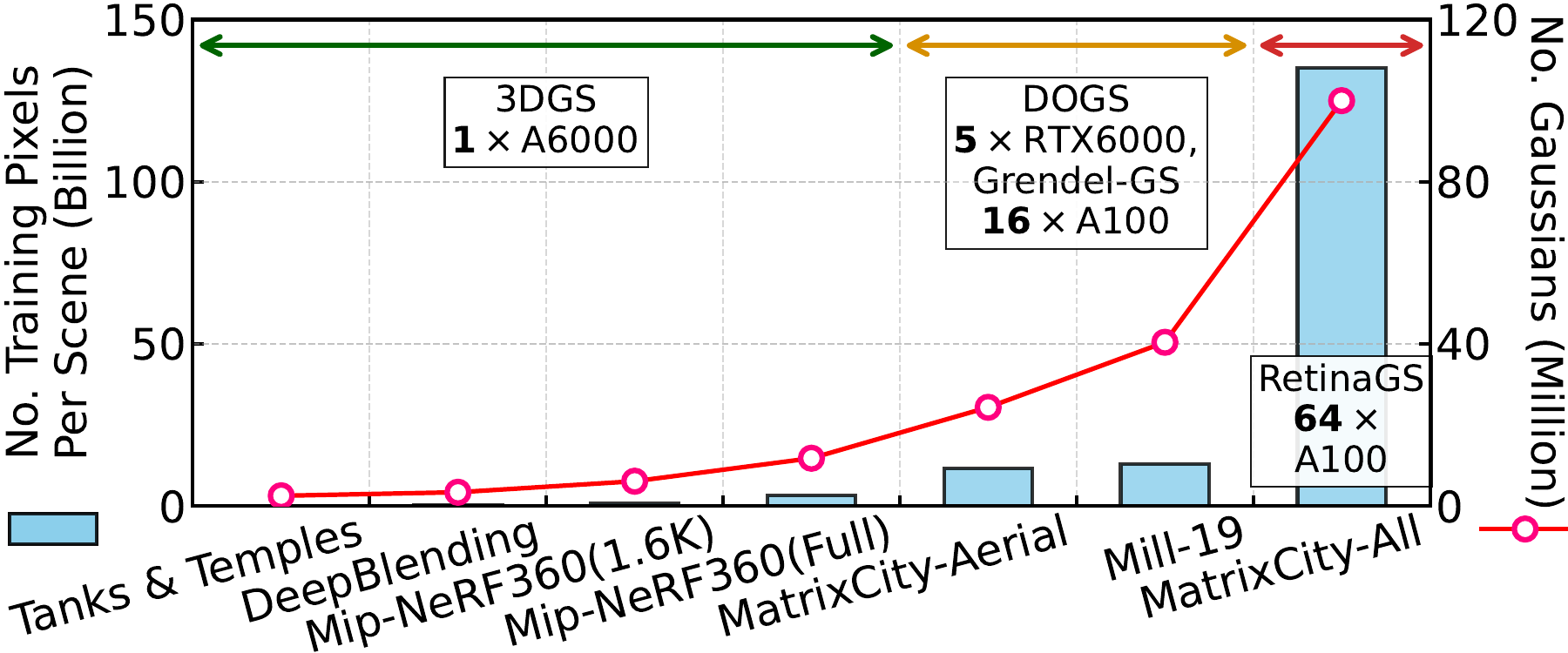}
    \caption{The trend of training \gaussian for larger, more detailed scenes challenges both software and hardware. As the number of training pixels grows, software has evolved from 3DGS~\cite{kerbl20233d} to distributed frameworks like Grendel-GS~\cite{grendel-gs} and RetinaGS~\cite{retinags}, while hardware has scaled from 1 GPU to 64 GPUs, with Gaussians per scene increasing by 20-100$\times$.}
    \label{fig:scaling}
\end{figure}

To facilitate the deployment of 3D-GS, prior works proposed compression pipelines to minimize the number of Gaussians~\cite{pup-3dgs, xie2025mesongs,fan2023lightgaussian, papantonakis2024reducing, compressed-3dgs} or create compact representations for \gaussian~\cite{chen2025hac, lu2024scaffold, navaneet2024compgs} while preserving rendering quality. Despite promising results, existing methods either modify the training procedure for compactness (i.e., \textbf{retraining}) or post-process a pre-trained model through fine-tuning (i.e., \textbf{refinement}), both of which involve training and pose new challenges to applicability: Both retraining and fine-tuning require the full training pipeline, including access to often limited training data and computing resources, and can be time-consuming.

More recently, researchers aim to handle high-resolution (e.g., 1080p and 4K) and large-scale 3D reconstruction tasks with tens of millions of Gaussians~\cite{chen2025dogs, grendel-gs, retinags}. However, given the huge number of Gaussians, 3D-GS cannot be trained nor rendered on a single GPU regardless of the batch size because it exceeds the single GPU memory capacity. 
As shown in \fref{fig:scaling}, while small to moderate-scale scenes, like those in Tanks\&Temples~\cite{knapitsch2017tanks} and Mip-NeRF360~\cite{mip-nerf360}, can be processed on a single Nvidia A6000, large-scale scenes like MatrixCity~\cite{li2023matrixcity} and Mill-19~\cite{turki2022megamill19} require distributed multi-GPU setups (e.g., 16-64) due to the additional memory demands of holding training images and optimizer states. The \textsc{Rubble} scene from Mill-19~\cite{turki2022megamill19}, trained using Grendel-GS~\cite{grendel-gs}, contains over 40 million Gaussians and a \fixme{8.9} GB file size with a large training memory footprint. Unfortunately, existing 3D-GS compression work often overlook the compute requirements associated with high-resolution and large-scale 3D construction tasks, which in fact has a big impact to the deployment of 3D-GS {with limited resources}.

As an initial attempt to address these challenges, FCGS~\cite{chen2024FCGS} introduces a generalizable model that compresses Gaussians without training, achieving state-of-the-art performance in under a minute. However, it falls short in two areas: (1) FCGS demands prohibitively high hardware resources, particularly memory (often over 24GiB), limiting its applicability and excluding resource-constrained mobile devices; (2) it lacks flexibility to adjust to rapidly varying compression needs, with each adjustment still taking about a minute.

To enable flexible and cost-efficient Gaussian compression, we introduce \name, a training-free method with low compute requirements (e.g., mobile deployable) and instant reconfigurability to adapt to varying compression needs. Specifically, unlike DNNs, where different layers exhibit diverse sensitivity, we observe that different Gaussian attribute channels exhibit different quantization sensitivity. As such, we propose to apply INT4/INT8 \emph{Channel-wise Mixed-Precision Quantization (MPQ)} scheme with configurable bit-widths across Gaussian attribute channels. In addition, different from previous approaches for compressing 3D-GS that rely on learning K-means-based codebooks~\cite{navaneet2023compact3d,compressed-3dgs}, which require iterative optimizations, our approach is non-iterative, which significantly reduces the compression overhead while minimizing the compression error.  
We further show that such channel-wise sensitivity can be exploited for semi-structured pruning as well, and we introduce \emph{Attribute-Discriminative Pruning (ADP)}, which enables fine-grained pruning at both the Gaussian and attribute granularities while eliminating the need to train the pruning mask~\cite{xie2025mesongs}, which greatly simplifies the pruning method for 3D-GS.  

\name exposes a broad optimization space to explore and identify the optimal compression rate within hardware-defined constraints. 
To navigate this space, we introduce \emph{Fast Online Adaption (FOA)}, a lightweight yet effective adaptation module that jointly considers \adp and \mpq to identify the optimal parameter combinations for meeting hardware-specified constraints, such as target quality or compression ratio. With no training involved, the time cost to reach each trade-off spot is reduced to seconds, which in turn enables \foa to quickly search multiple spots.

Overall, in response to the rising costs of large-scale Gaussian training and varying compression needs, \name  eliminates the need for training in \gaussian compression. This approach avoids the high-end hardware requirements incurred by storing training data and optimizer states on GPUs, reduces compression time costs, and offers flexibility in adjusting compression rates and qualities.

We conduct extensive experiments on both desktop and resource-constrained mobile platforms across diverse datasets to compress \gaussian models for novel view synthesis, covering scene scales and image resolutions from small to large. Our evaluation shows that \name achieves a significant compression ratio while maintaining high rendering quality, e.g., a \fixme{96.4\%} model size reduction with $<$1 dB drop in PSNR, with compression rates fully adjustable at negligible cost. Compared to state-of-the-art methods, \name is \fixme{10-100$\times$} faster than training-involved methods and \fixme{1.7-2.1$\times$} faster than the training-free FCGS~\cite{chen2024FCGS}. It also adapts rapidly to varying memory or bandwidth constraints, with each adjustment taking just 1 to 2 seconds, making it a flexible and cost-effective solution for Gaussian compression.

\section{Related Work}

\noindent
\textbf{3D Gaussian compression}.
Many studies focus on 3D-GS compression. EAGLES~\cite{girish2023eagles} introduces an encoder-decoder network to compress Gaussian attributes into a latency code. Scaffold-GS~\cite{lu2024scaffold} uses anchor points to guide the distribution of 3D Gaussians and learns to predict their attributes at these points. LightGaussian~\cite{fan2023lightgaussian} distills Spherical Harmonics(SHs) to a lower degree. 
Compressed3D~\cite{compressed-3dgs} introduces sensitivity-aware clustering, quantization-aware fine-tuning, and entropy encoding to reduce \gaussian size.  
Similarly, RDO-Gaussian~\cite{rdogaussian} learns adaptive pruning of SHs and assigns varying SH degrees based on material and illumination needs. 
While promising, these methods all require significant training effort to identify redundancies and condense 3D Gaussian information. FCGS~\cite{chen2024FCGS} develops a universal model for compression without training, but still suffers from insufficient memory even on GPUs like RTX 3090 and A6000. This hinders their applicability in resource-constrained scenarios, such as mobile devices or large-scale 3D-GS that requires expensive multi-GPU setups for training~\cite{grendel-gs,retinags}. 
Our work differs by introducing a flexible, cost-effective compression method that adaptively adjusts compression levels while maintaining comparable quality, with lower resource usage and suitability for even resource-constrained mobile devices.

\noindent
\textbf{Slimmable scene representations.}
Training a single model that allows for trade-offs between task quality and model size at runtime is crucial, especially given the varying resources available. This model can be classical Deep Neural Networks~\cite{yu2018slimmable, yu2019universally, li2021dynamic, hou2021slimmable} or recent scene representations such as NeRF~\cite{mildenhall2021nerf} or 3D Gaussians~\cite{kerbl20233d} for rendering tasks.
Recent research achieves slimmability in NeRF representations. CCNeRF~\cite{tang2022compressible} uses rank truncation to dynamically control model size and levels of detail through hybrid tensor rank decomposition. Similarly, SlimmeRF~\cite{yuan2024slimmerf} also employs rank truncation but incrementally builds the rank during training. While there has been little prior work on achieving slimmability in 3D Gaussians, our work presents a hybrid method that dynamically reduces model size while maintaining quality, which cannot be accomplished by simply applying individual designs.

\section{Challenges in Training-Free Compression}
\label{sec:prelim}

Post-training compression methods have demonstrated great efficiency in deep learning (DL) and large language model (LLM) compression, as they are applied without retraining or fine-tuning~\cite{gptq,awq,smooth-quant}. Among these strategies, two commonly used ones are {quantization} and {pruning}. However, there is few investigation into \fixme{training-free compression} for 3D-GS. Here, we briefly discuss the challenges of applying post-training quantization and pruning to 3D-GS.

\begin{figure}[t!]
    \centering
    \captionsetup[subfigure]{width=0.5\columnwidth}
    \subfloat[\small{RTN with varying bit-widths.}]
    {
        \includegraphics[width=.48\columnwidth]{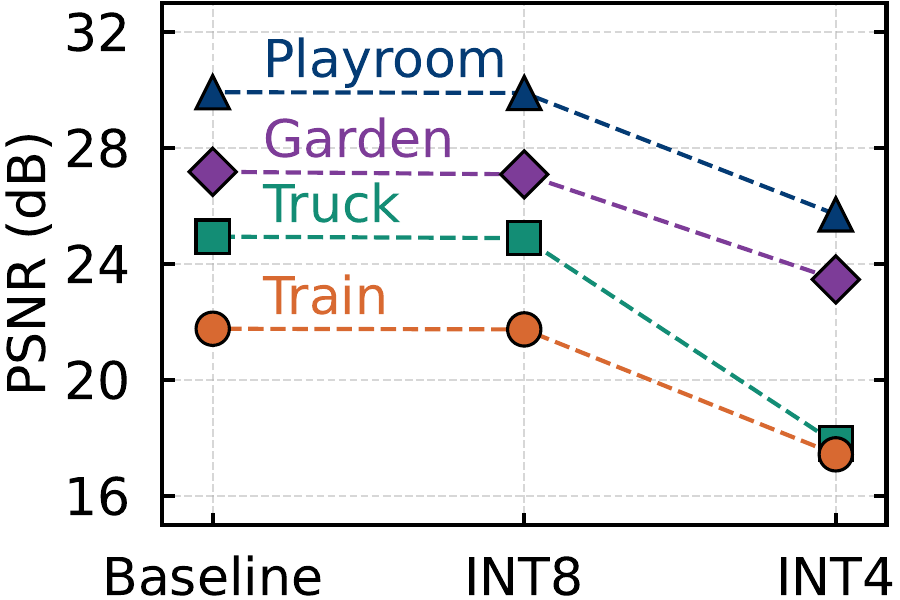}
        \label{fig:sense-scene}
    }
    \subfloat[\small{Varying ranges in attributes.}]
    {
        \includegraphics[width=.48\columnwidth]{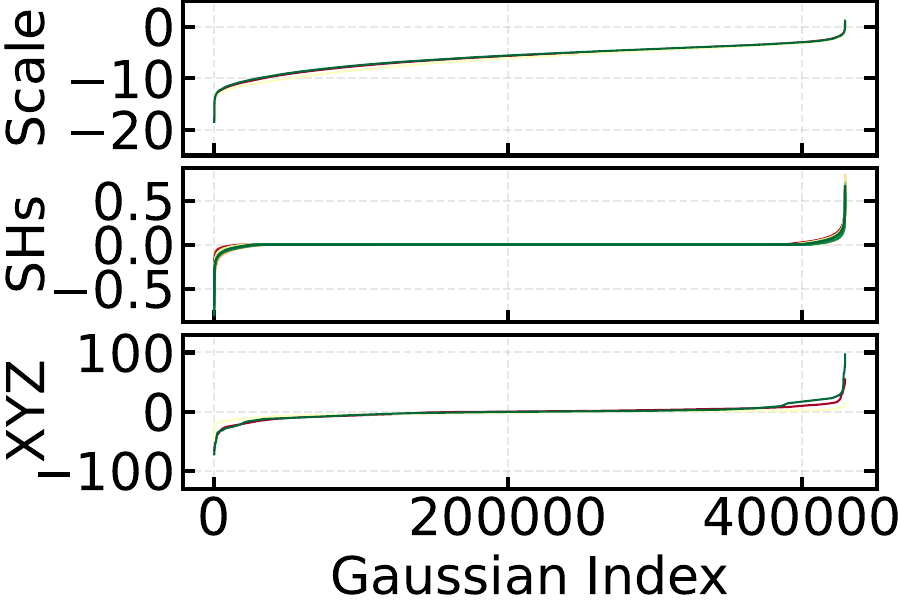}
        \label{fig:dynamic}
    }
    \caption{Left: Quality of quantized \gaussian with varying attribute bit-widths across scenes. Right: \gaussian attributes show divergent ranges, with up to 200$\times$ difference between XYZ to SHs.}
    \label{fig:sense}
\end{figure}

\noindent
\textbf{Non-negligible accuracy loss with ultra-low quantization.}  For a 3D-GS model with $N\times M$ parameters, existing post-training quantization schemes~\cite{awq,gptq}, such as Round-to-Nearest (RTN), reduce the storage by four times compared to FP32 by quantizing the parameter values to INT8. However, according to \fref{fig:sense-scene}, while 8-bit quantization largely maintains rendering quality across scenes, further reducing it to ultra-low bits, e.g., 4-bit, consistently lead to a sharp performance drop in rendered scenes. 

\begin{figure*}[ht!]
    \centering
    \includegraphics[width=2.1\columnwidth]
    {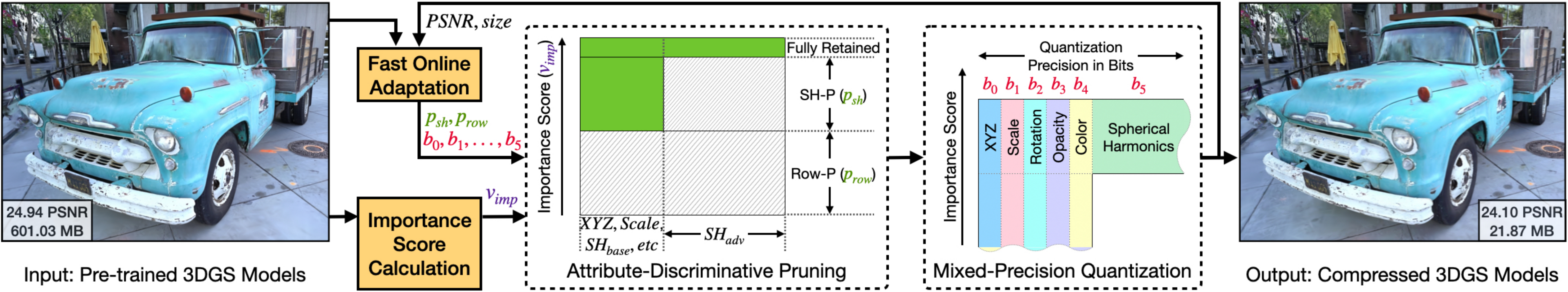}
    \caption{Overview of \name. \name first applies attribute-discriminative pruning (ADP) to obtain semi-structurally pruned Gaussian primitives. It then employs INT8/INT4 channel-wise mixed-precision quantization (MPQ) to further reduce the model size along the bit dimension. Finally, \name introduces a novel lightweight online adaptation (FOA) algorithm to adaptively adjust the compression ratio for diverse scenes on hardware with different efficiency constraints.}
    \label{fig:overview}
\end{figure*}

\noindent
\textbf{Divergent attribute range.} To investigate why INT4 quantization leads to significant rendering quality drop, we analyze the per-attribute channel sensitivity both qualitatively and quantitatively. \fref{fig:dynamic} shows that different attribute channels have dramatically different ranges (more details in \sref{subsec:quantization}), and quantizing certain channels (e.g., scale) to ultra-low bits leads to large PSNR drop. \emph{This large variance in the value range makes it difficult to use a fixed quantization range (usually the maximum value) for all channels while preserving rendering quality}, as the limited representation power for large-range channels leads to large quantization errors.   

\noindent
\textbf{Performance degradation from zero-shot Gaussian pruning.} Prior work prunes 3D-GS by removing the least significant 3D Gaussians based on a calculated importance score for each Gaussian. 
Common importance metrics include Opacity-Based Pruning~\cite{gs-pruning-opacity, lp-3dgs}, which removes low-opacity Gaussian primitives due to their limited impact on visual quality; Magnitude-Based Pruning~\cite{lin2025metasapiens}, which prunes primitives with fewer ray intersections as they contribute less to accuracy; and Volumetric-Based Pruning~\cite{papantonakis2024reducing}, which evaluates primitives by their 3D volume, with larger volumes indicating greater significance for rendering quality. And Global Significance Score~\cite{fan2023lightgaussian}, which combines Gaussian opacity, 3D volume, and pixel hit magnitude across training views to quantify its overall contribution to visual quality. However, prior work finds that directly pruning Gaussians (e.g., zero-shot pruning based on the magnitude of their opacity values or second-order approximation of the reconstruction error on the training views) leads to significant quality degradation~\cite{fan2023lightgaussian,pup-3dgs}. This is why existing methods typically employ retraining, fine-tuning, or knowledge distillation to adjust the remaining Gaussians and minimize quality loss~\cite{papantonakis2024reducing, fan2023lightgaussian, rdogaussian, lu2024scaffold, chen2025hac, girish2023eagles, lee2024compact, navaneet2023compact3d, xie2025mesongs}. 
However, as the compute requirement for scaling 3D-GS is growing, a lightweight and efficient method is desirable for compressing 3D-GS. 

\section{Methodology}
\label{sec:design}

\subsection{Problem Formulation}
\label{subsec:problem}

The objective is to develop a lightweight and cost-effective method for training-free compression of 3D Gaussian models. Formally, the input of the problem includes an optimized 3D Gaussian model with a data layout of an $N \times M$ matrix, where each row ($N$) represents an individual 3D Gaussian, and each column ($M$) corresponds to an attribute channel. These channels include: ellipsoid's center coordinates (e.g., 3),  scaling factor along each dimension (e.g., 3), rotation (encoded in quaternion:  $w, x, y, z$), opacity, view-independent base color in RGB ($SH_{base}$), and 3-degree spherical harmonics that enable view-dependent effects ($SH_{adv}$). 
Our goal is to obtain a \emph{matching compressed 3D-GS model}, where its rendering quality is no less than $\epsilon$ (compression tolerance regime) in comparison with the uncompressed model. 
The solution should not require the training pipeline or access to the training images.

\subsection{\name Algorithm}

\name is flexible in compressing pre-trained 3D Gaussian models at negligible cost, to either achieve a target compression ratio or to meet a quality target. \fref{fig:overview} highlights the key components of the system and demonstrates this procedure on the \textsc{Truck} scene from the \texttt{Tanks\&Temples} dataset, achieving an \fixme{96.4\%} reduction in model size with $<$\fixme{1 dB} PSNR drop in just \fixme{20} seconds.

\subsubsection{INT4/INT8 Channel-wise Mixed-Precision Quantization (MPQ)}
\label{subsec:quantization}

Prior work on DNN compression find that different DNN layers exhibit diverse sensitivity~\cite{wang2019haq, dong2019hawq, mpqdiffnas}, which motivates mixed-precision quantization, where more bits are assigned to sensitive layers to preserve model quality. Recent work has also explored spatial sensitivity in 3D-GS models by computing a second-order Hessian over a hypothetical perturbation to model weights and using the sensitivity score for pruning~\cite{pup-3dgs}. While being effective, it calculates the second order approximation for individual Gaussian, which is per-Gaussian and very expensive to calculate. 

\begin{figure}[ht!]
  \begin{minipage}[t]{0.48\columnwidth}
    \centering
    \includegraphics[width=\columnwidth]{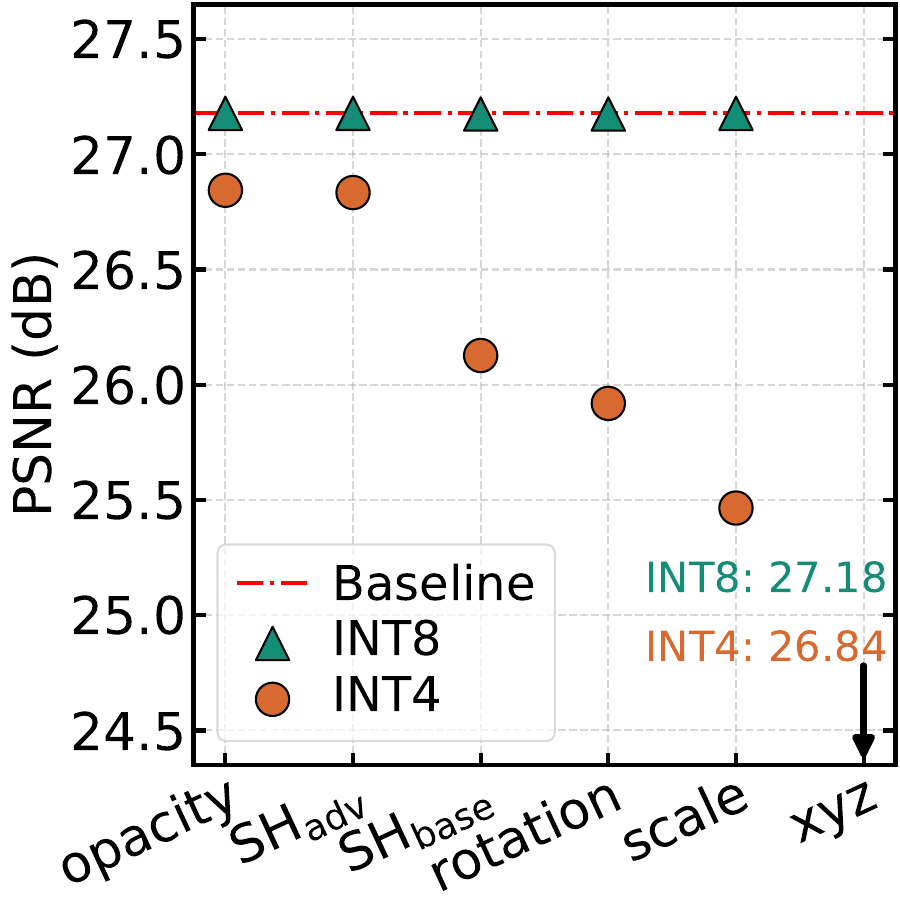}
    \caption{Attribute channels exhibit divergent impact to rendering quality, evidencing the need for mixed-precision quantization that accounts for attribute sensitivity.}
    \label{fig:sense-channel}
  \end{minipage}
  \hfill
  \begin{minipage}[t]{0.48\columnwidth}
    \centering
    \includegraphics[width=\columnwidth]{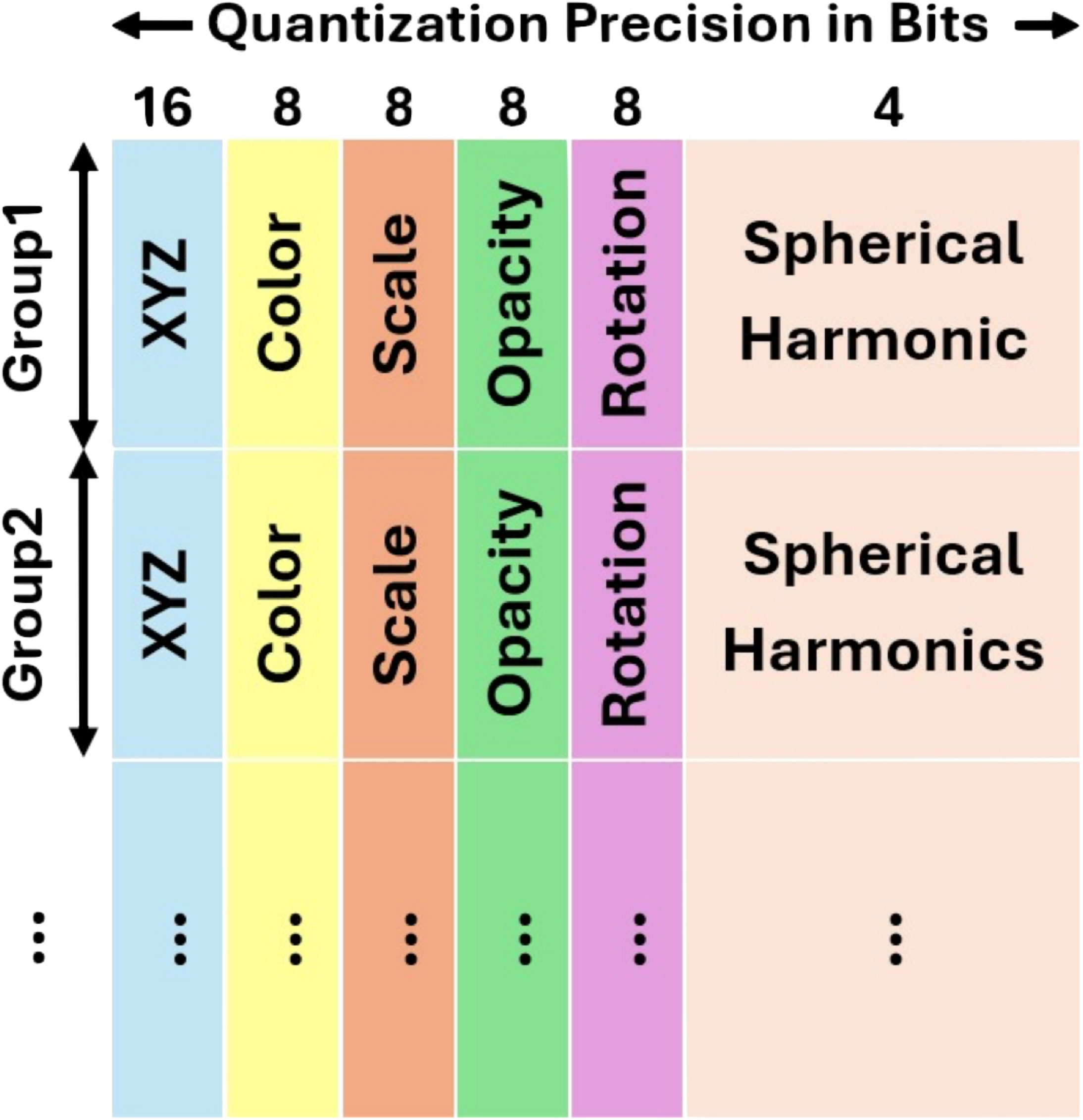}
    \caption{Channel-wise mixed-precision quantization. The same color denotes the same bit-width. Group \# indicates sub-groups with their own quantization ranges.}
    \label{fig:quantization_grouped}
  \end{minipage}
\end{figure}

Intuitively, different attribute channels store distinct information (e.g., geometric, textual); do they exhibit varying quantization sensitivity? Our analysis in \fref{fig:sense-channel} shows the quality impact of applying 4-bit and 8-bit quantization to each attribute channel. The results show that channels in 3D-GS have varying sensitivities, and assigning the same number of bits to all channels is sub-optimal.

Based on the observation, we propose the \emph{channel-wise mixed-precision quantization} method: We apply an ultra-low quantization bits (e.g., INT4) to each individual channel separately and keep the other channels in INT8 to find a normalized loss gap value between the uncompressed and quantized model. We use INT4 on channels with smaller gaps. While this approach does not explicitly consider the correlation between channels, we find that it is robust and efficient to select less sensitive columns. In addition, we observe minimal variation in channel sensitivity to bit-widths across different scenes, as certain Gaussian attributes (e.g., positions) are more prone to inaccuracies than others. Therefore, we use the same set of channel-specific bit-widths but can support the search procedure with minimal cost (several seconds) when necessary.

While selective channel-wise mixed-precision quantization greatly reduces quantization error, we find that directly quantizing an entire channel with the same quantization range still leads to non-trivial accuracy degradation, as channels can contain millions of elements in some scenes. To tackle this, we use subchannel-wise grouped quantization for 3D-GS models. In each channel, we bucket sequential attributes together as sub-groups, e.g., all attributes of each set of channels are arranged to 1000 sub-groups. Each sub-group has its own quantization range, as shown in \fref{fig:quantization_grouped}. Note that this method does not require iterative optimizations like prior methods~\cite{navaneet2023compact3d,compressed-3dgs}, which is hard to scale with increasing Gaussian primitives.

\subsubsection{Attribute-Discriminative Pruning (ADP)}
Prior works effectively prune the number of Gaussian primitives but either overlook fine-grained pruning opportunities along the attribute dimension~\cite{fan2023lightgaussian} or require iterative training to learn the pruning mask~\cite{xie2025mesongs}. Our observation is that geometry attributes, such as position, rotation, and scale, have a greater impact on quality with a smaller data size (e.g., 11 vs. 59 in this case). In contrast, texture attributes such as $SH_{adv}$ require more storage (e.g., 45 attributes per Gaussian) but have less influence on quality (\fref{fig:sense-channel}).

\begin{figure}[t!]
  \begin{minipage}[t]{0.48\columnwidth}
    \centering
    \includegraphics[width=\columnwidth]{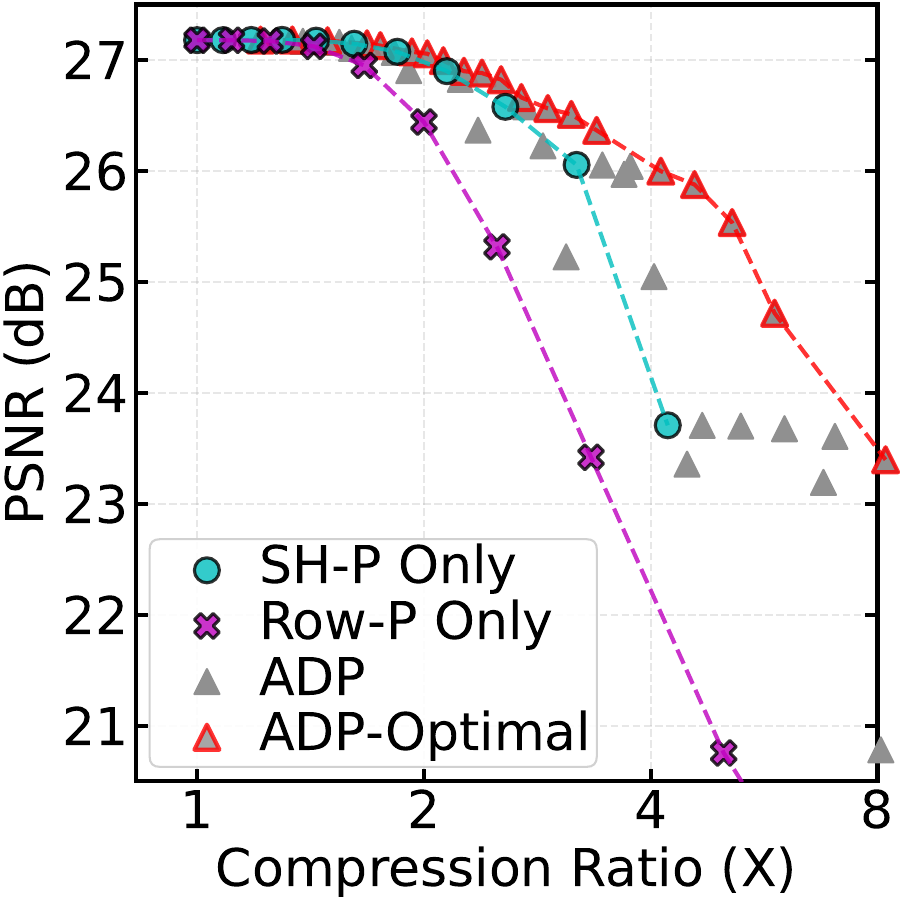}
    \caption{A broad space for balancing quality and compression ratio, with optimal designs (ADP Optimal) outperforming pruning methods that ignore attribute sensitivity (Row-P and SH-P).}
    \label{fig:mixed-dim}
  \end{minipage}
  \hfill
  \begin{minipage}[t]{0.48\columnwidth}
   \centering
    \includegraphics[width=\columnwidth]{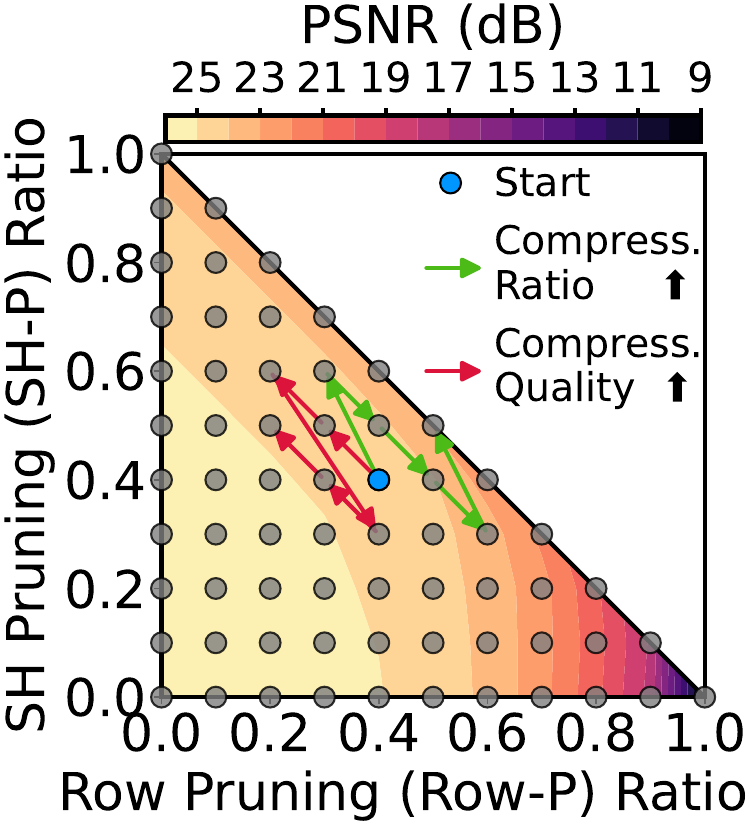}
    \caption{\foa operates within the \adp design space, with optimal targets near the diagonal under quality constraints. Path with green arrows prioritizes compression, and red arrows favor quality.}
    \label{fig:design-space}
  \end{minipage}
\end{figure}

We exploit this opportunity by proposing \pruning, a semi-structured pruning method that effectively leverages the attribute-level importance properties to perform pruning.
In this work, we consider a specific importance score from LightGaussian~\cite{fan2023lightgaussian} -- the global significant score of each Gaussian. In particular, the score $GS_n$ at the $n-$th row is calculated as:
\begin{equation}
    GS_n = \sum_{n=1}^{MHW}\mathbbm{1}(G(X_n), r_i)\cdot \sigma_n \cdot \gamma(\bigoplus_n)
\end{equation}
where $i$ represents a pixel, MHW represents the number of training views, image height, and width, respectively. $\mathbbm{1}$ is the indicator function that determines whether a Gaussian intersects with a given ray $r_i$. $\sigma_n$ is the Gaussian's opacity, and $\gamma(\bigoplus_n)$ measure the 3D Gaussians' volume based on the scaling factors. The metric jointly considers how each Gaussian contributes to the global pixels, their opacity, and its volumetric density, which has been shown to be effective in pruning Gaussians. Note that the global significance score only needs to be calculated once given that the Gaussian attributes are fixed after training.

Instead of fully pruning or retaining Gaussian primitives, \adp enables partial pruning, where less critical attributes, such as $SH_{adv}$, can be partially pruned from non-critical Gaussians. \fref{fig:overview} conceptually illustrates how \adp prunes Gaussians along both the primitive and attribute dimensions, with shaded areas representing the pruned parts. By sorting Gaussians based on importance scores in descending order, \adp retains the top $\alpha\%$ of Gaussians and discards the bottom $\beta\%$ that contribute minimally to quality. Unlike prior methods~\cite{fan2023lightgaussian, xie2025mesongs, navaneet2024compgs}, \adp partially retains the remaining Gaussians (1 - $\alpha\%$ - $\beta\%$), preserving only critical attributes without re-training.

\adp has two parameters: the ratio of Gaussians pruned along the entire row (Row-P, i.e., $\alpha\%$) or partially for $SH_{adv}$ (SH-P, i.e., 1 - $\alpha\%$ - $\beta\%$). Their combination creates trade-offs between quality and compression ratio, as shown in \fref{fig:mixed-dim}. Notably, a Pareto-optimal frontier is formed with designs enabled by \adp, whereas simply applying Row-P as in prior work~\cite{fan2023lightgaussian, pup-3dgs} or SH-P individually leads to sub-optimal compression performance. \adp exposes a large set of parameter candidates, enabling flexible compression priorities.

\newcolumntype{?}{!{\vrule width 2pt}}

\begin{table*}[t!]
	\caption{
		\label{tab:quant}{Quantitative comparison of quality scores, model sizes, and time costs for \name and prior \gaussian compression methods, measured on our desktop with a single RTX 3090 GPU. File size is in \underline{MiB}, and time cost is in \underline{Seconds}. Refinement-based, retraining-based, and training-free methods are shown in \colorbox{olive!10}{yellow}, \colorbox{gray!20}{gray}, and \colorbox{pink!20}{pink}, respectively, with \gaussian in \colorbox{blue!5}{blue} as a reference. The best, second-best, and third-best methods in file size and time cost are shown in \best{green}, \second{cyan}, and \third{orange}.}
	}
    \small
    \centering
	\scalebox{1.035}
    {
        \begin{threeparttable}
        \setlength{\tabcolsep}{2pt}
		\begin{tabular}{l|ccccc|ccccc|ccccc}
            \toprule[0.15em]

        	Dataset & \multicolumn{5}{c|}{\textbf{Mip-NeRF360}}  & \multicolumn{5}{c|}{\textbf{Tanks\&Temples}} & \multicolumn{5}{c}{\textbf{Deep Blending}} \\
        	Method $\mid$ Metric
        	& PSNR$^\uparrow$   & SSIM$^\uparrow$    & LPIPS$^\downarrow$  & Size$^\downarrow$ & Time$^\downarrow$ & PSNR$^\uparrow$   & SSIM$^\uparrow$    & LPIPS$^\downarrow$  & Size $^\downarrow$ & Time$^\downarrow$ & $PSNR^\uparrow$   & $SSIM^\uparrow$    & $LPIPS^\downarrow$  & Size$^\downarrow$ & Time$^\downarrow$ \\
        	\midrule[0.05em]
        	\cellcolor{blue!5} 3D-GS~\cite{kerbl20233d} &
        	27.21 & 0.815 & 0.214 & 795.26 & 1576.25 & 23.14 & 0.841 & 0.183 & 421.91 & 963.87 & 29.41 & 0.903 & 0.243 & 703.77 & 1654.11 \\
        	\cellcolor{olive!10} LightGaussian~\cite{fan2023lightgaussian} &
        	26.96 & 0.800 & 0.244 & 52.05  & 917.61 & 23.14 & 0.818 & 0.222 & 27.87 & 426.18 & 28.82 & 0.891 & 0.273 & 45.85 & 686.04 \\
        	\cellcolor{olive!10} Compressed3D~\cite{compressed-3dgs} &
        	27.02 & 0.803 & 0.236 & \second{28.81} & \third{272.84} & 23.34 & 0.836 & 0.192 & \third{17.25} & 191.45 & 29.44 & 0.902 & 0.250 & \second{25.28} & \third{247.98} \\
        	\cellcolor{olive!10} PUP 3DGS~\cite{pup-3dgs} &
        	26.66 & 0.789 & 0.267 & 79.53 & 427.99 & 22.46 & 0.799 & 0.248 & 42.19 & \third{189.69} & 29.40 & 0.903 & 0.256 & 70.38 & 270.76 \\
        	\cellcolor{gray!20} CompGS~\cite{navaneet2024compgs} &
        	26.99 & 0.801 & 0.250 & \best{21.08} & 2451.07 & 23.21 & 0.837 & 0.201 & \best{14.20} & 1397.70 & 29.98 & 0.911 & 0.250 & \best{15.15} & 1941.88 \\
        	\cellcolor{pink!20} FCGS-Raw~\cite{chen2024FCGS}\tnote{\small$\ast$} &
        	--- & --- & --- & --- & --- & --- & --- & --- & --- & --- & --- & --- & --- & --- & --- \\
            \cellcolor{pink!20} FCGS-Opt\tnote{\small$\ddagger$} &
        	27.04 & 0.799 & 0.231 & 62.34 & \second{53.57} & 23.36 & 0.839 & 0.186 & 31.12 & \second{30.09} & 29.61 & 0.902 & 0.243 & 55.41 & \second{42.61} \\
        	\midrule[0.05em]
            \cellcolor{pink!20} \name(Ours) &
        	26.38 & 0.780 & 0.251 & \third{40.80} & \best{25.69} & 22.44 & 0.804 & 0.219 & \second{16.30} & \best{18.24} & 28.61 & 0.884 & 0.269 & \third{25.48} & \best{25.65} \\

            \bottomrule[0.15em]
        \end{tabular}
        
        \begin{tablenotes}\footnotesize
            \item[{$\small\ast$}] FCGS-Raw is the official implementation. It runs out of the 24 GiB VRAM of a single RTX 3090 on 7 of the 13 scenes. See per-scene results and discussions in \fixme{\tref{tab:per-scene-all}}.
            \item[\small{$\ddagger$}] FCGS-Opt is our customized version with reduced memory usage and faster speed, without compromising compression ratio or quality.
        \end{tablenotes}
    \end{threeparttable}
	}
\end{table*}

\subsubsection{Fast Online Adaptation (FOA)}
\label{subsec:adaptation}

While identifying a single configuration for \adp and \mpq that generalizes across scenes is challenging, we find that the quality impact of \mpq is scene-insensitive, mainly due to its scene-agnostic operation, while the primary sensitivity arises from the importance score calculation for \adp. We extend the analysis in \fref{fig:mixed-dim} to three additional scenes from different datasets and observe that, although the quality drop varies, the linear correlation between quality loss and pruning ratio remains consistent across scenes. Therefore, the Pareto-optimal frontier shares the same set of parameter pairs, narrowing down the optimal parameter configurations to a limited set of candidates.

We introduce a fast online adaptation (\foa) module that efficiently searches for the optimal design based on input scenes at negligible cost. Given a user-specified target quality or compression ratio, \foa searches in the candidate sets until reaches the best compression performance in observance to the constraints. While the search naturally can be done in parallel since each is independent to others, we exploit the fact that the Pareto-optimal frontier is convex, so that the priority trend at each direction is monotonious.

\foa begins with a one-time cost for model I/O and importance score calculation. At each search step, \foa duplicates the input Gaussians, then prunes, quantizes, dequantizes, and evaluates quality using standard metrics like PSNR, repeating until the best compression parameters are found. Each search yields different quality-size trade-offs within seconds, enabling instant reconfigurability, which is impractical for training-involved methods.

\fref{fig:design-space} illustrates the \foa procedure. The search aims to approach the diagonal (representing higher compression ratios) within the colorized zones (i.e., quality constraint). Starting from the initial state (blue), the search can traverse either along the green direction, prioritizing compression ratio, or the red direction, prioritizing quality. The traversal can occur bidirectionally when starting from the middle of the path, typically completing within a few seconds.

\section{Evaluation}
\label{sec:eval}

\subsection{Evaluation Methodology}

\textbf{Datasets.}
We follow standard practices~\cite{kerbl20233d, fan2023lightgaussian} to evaluate \name on three datasets: Mip-NeRF360~\cite{mip-nerf360}, Tanks\&Temples~\cite{knapitsch2017tanks}, and Deep Blending~\cite{hedman2018deep}, using 9, 2, and 2 scenes, respectively. These datasets span a wide variety of scenes, including both bounded indoor and unbounded outdoor scenarios, with diverse capture styles, object distributions, and levels of detail.

\textbf{Baselines.}
We compare \name with 3DGS~\cite{kerbl20233d} and several compression methods. We compare with FCGS~\cite{chen2024FCGS}, the only known training-free method recently introduced. For refinement-based baselines, we select LightGaussian~\cite{fan2023lightgaussian}, Compressed3D~\cite{compressed-3dgs}, and PUP-3DGS~\cite{pup-3dgs}, and for retraining-based methods, we use CompGS~\cite{navaneet2024compgs}, collectively referred to as \textbf{training-involved methods}. Our training-free design enables instant adaptation to various compression targets at minimal cost. In our main experiments, we target a $<$1 dB PSNR drop and evaluate adaptability separately.

Note that FCGS fails on 7 of the 13 test scenes due to excessive memory usage, exceeding the 24 GiB VRAM of the RTX 3090 used in our setup.  We therefore compare with a customized variant, which reduces memory usage and time cost while preserving the original compression ratios and quality. Details are provided in \tref{tab:per-scene-all}.

\textbf{Metrics.}
We evaluate compression ratio, quality, and time cost. The compression ratio is calculated as the file size of the \gaussian models divided by the uncompressed baseline. Compression quality is assessed using peak signal-to-noise ratio (PSNR), structural similarity (SSIM)~\cite{wang2004image}, and perceptual similarity (LPIPS)~\cite{zhang2018unreasonable}, by comparing images rendered from models before and after compression. Time cost refers to the total time required for executing the compression methods, either the post-processing time for refinement-based methods or the total training time for retraining-based methods.

\textbf{Hardware setups.}
{Experiments are primarily conducted on a setup with an Intel Core i9-10900K, 64 GB of DRAM, and an Nvidia RTX 3090 GPU with 24 GB VRAM, unless otherwise noted. We also deploy our training-free design on the Nvidia Jetson Xavier~\cite{Jetson_xavier_agx}, a mobile platform with 16 GB of shared memory between the CPU and a 512-core Volta GPU, to assess its lightweight advantage. To obtain large \gaussian models, we train Grendel-GS~\cite{grendel-gs} on 4 A100 GPUs, each with 40 GB of VRAM, interconnected via NVLink.}

\newcommand{\zoominwithcomment}[9]{

  \begin{tikzpicture}[spy using outlines={rectangle,#8,magnification=3,size=#6}]
    \node[anchor=south west, inner sep=0] (img) {
      \includegraphics[width=#7]{#1}
    };

    \spy on (#2, #3) in node at (#4,#5);

    \node[anchor=south east, xshift=-2pt, yshift=2pt, text=white]
      at (img.south east) {\small #9};
  \end{tikzpicture}
}
\begin{figure*}
	\newlength\mytmplen
	\setlength\mytmplen{.193\linewidth}
	\setlength{\tabcolsep}{1pt}
	\renewcommand{\arraystretch}{0.5}
	\centering
	\begin{tabular}{cccccc}
		Ground Truth& Ours \raisebox{-0.4ex}{\small (single/avg)} & 3DGS \raisebox{-0.4ex}{\small (single/avg)} & Compressed3D \raisebox{-0.4ex}{\small (single/avg)} & CompGS \raisebox{-0.4ex}{\small (single/avg)} \\
        
		\zoomin{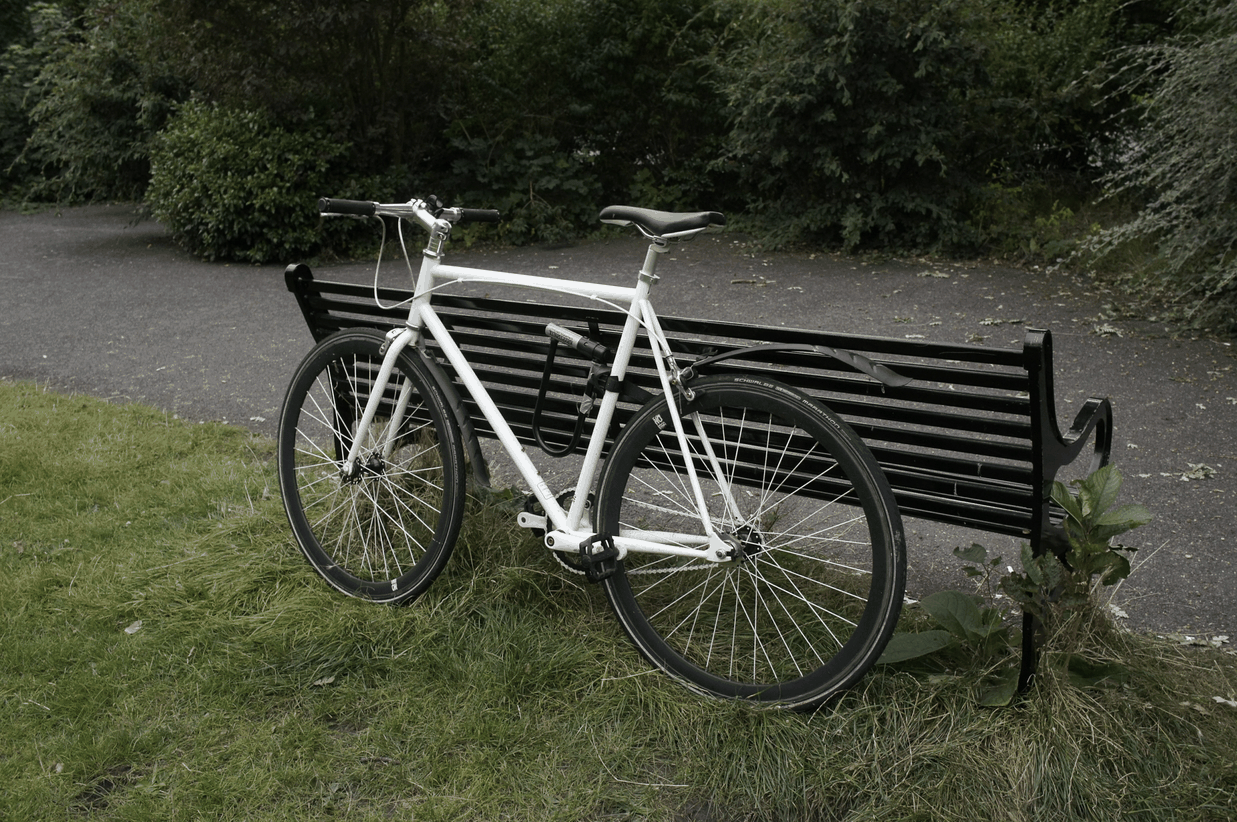}{3.0}{1.1}{0.5 5cm}{0.55cm}{1cm}{\mytmplen}{3}{red}&
        
        \zoominwithcomment{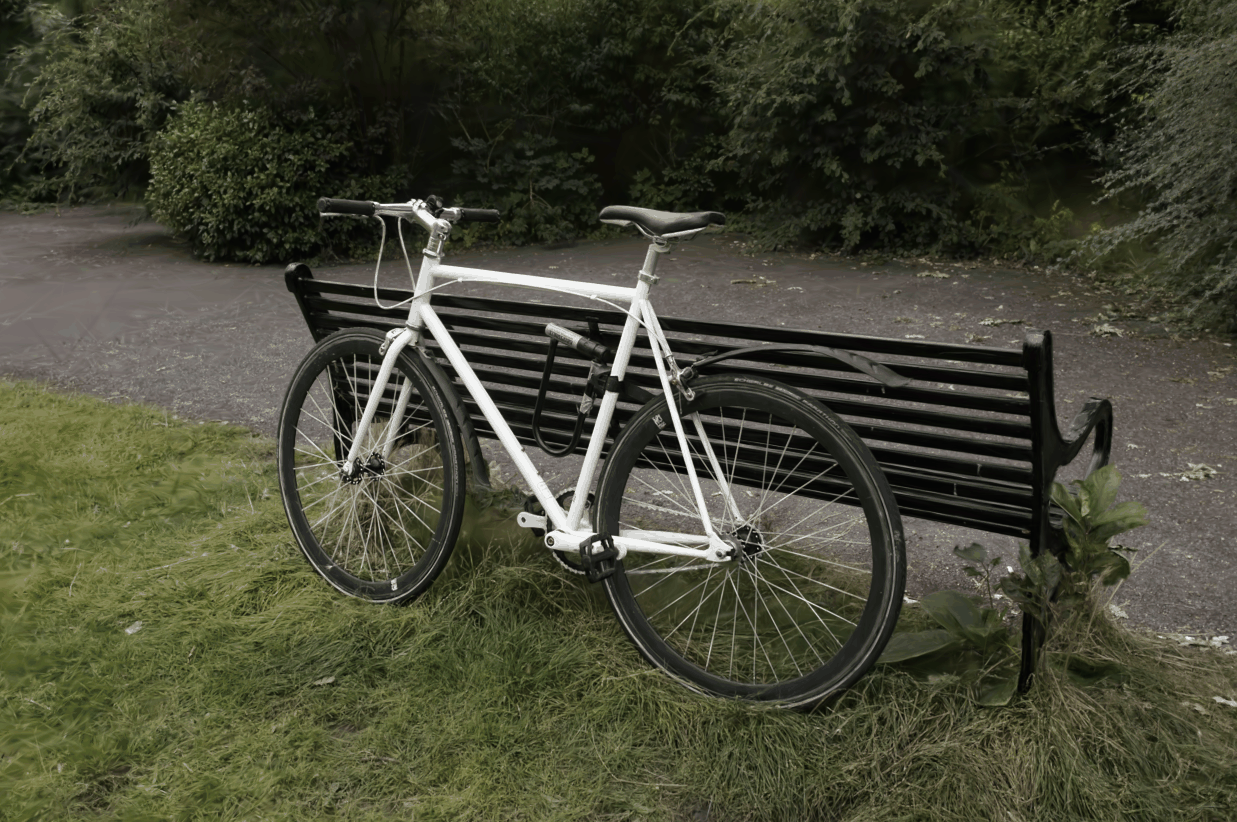}{3.0}{1.1}{0.55cm}{0.55cm}{1cm}{\mytmplen}{red}{25.37 / 24.17}&
        
        \zoominwithcomment{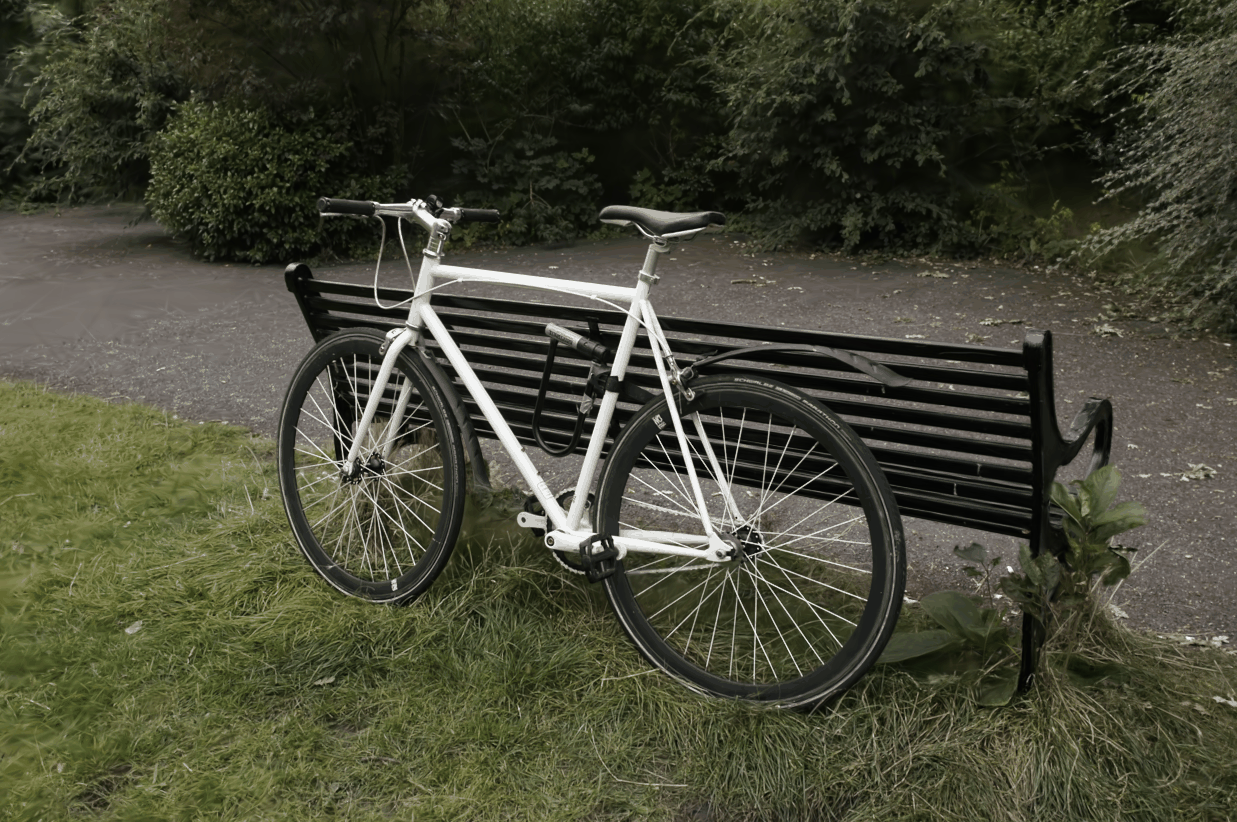}{3.0}{1.1}{0.55cm}{0.55cm}{1cm}{\mytmplen}{red}{26.58 / 25.17}&
        \zoominwithcomment{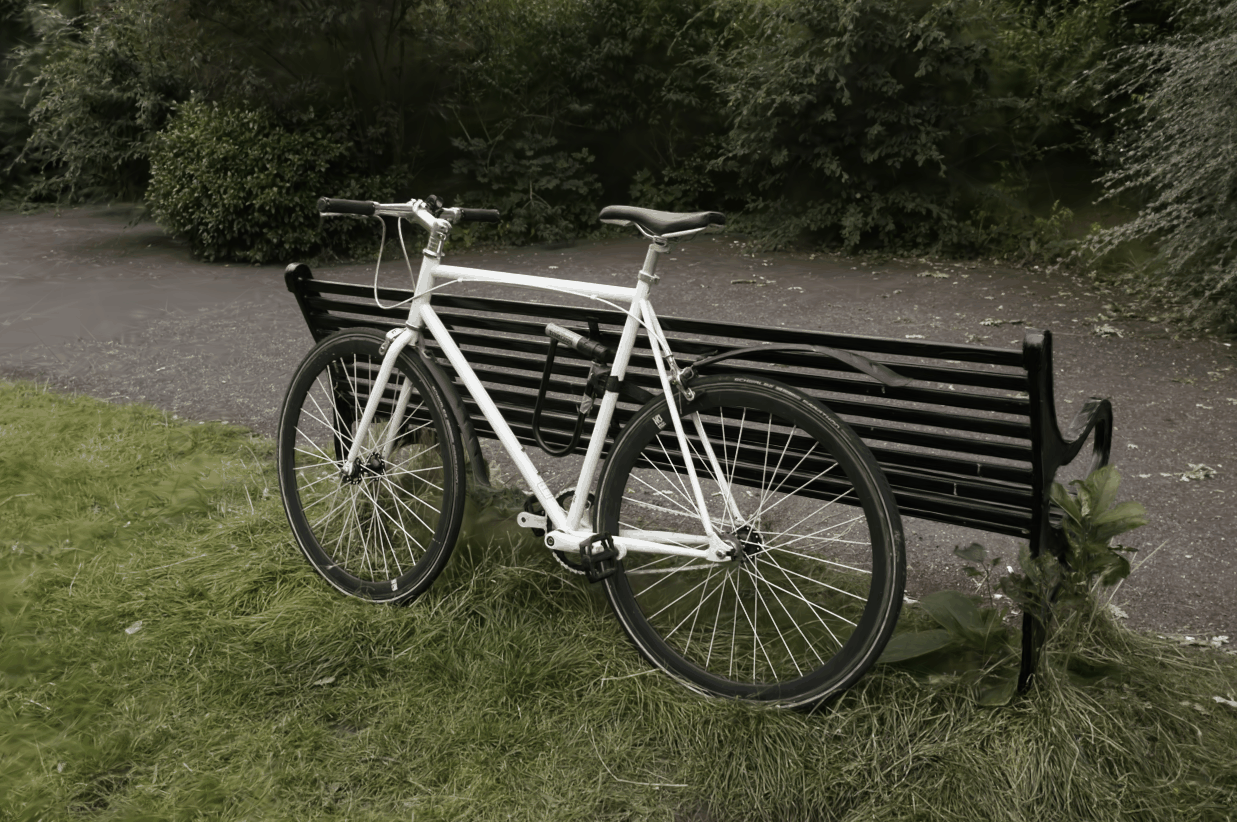}{3.0}{1.1}{0.55cm}{0.55cm}{1cm}{\mytmplen}{red}{26.22 / 24.97}&
        \zoominwithcomment{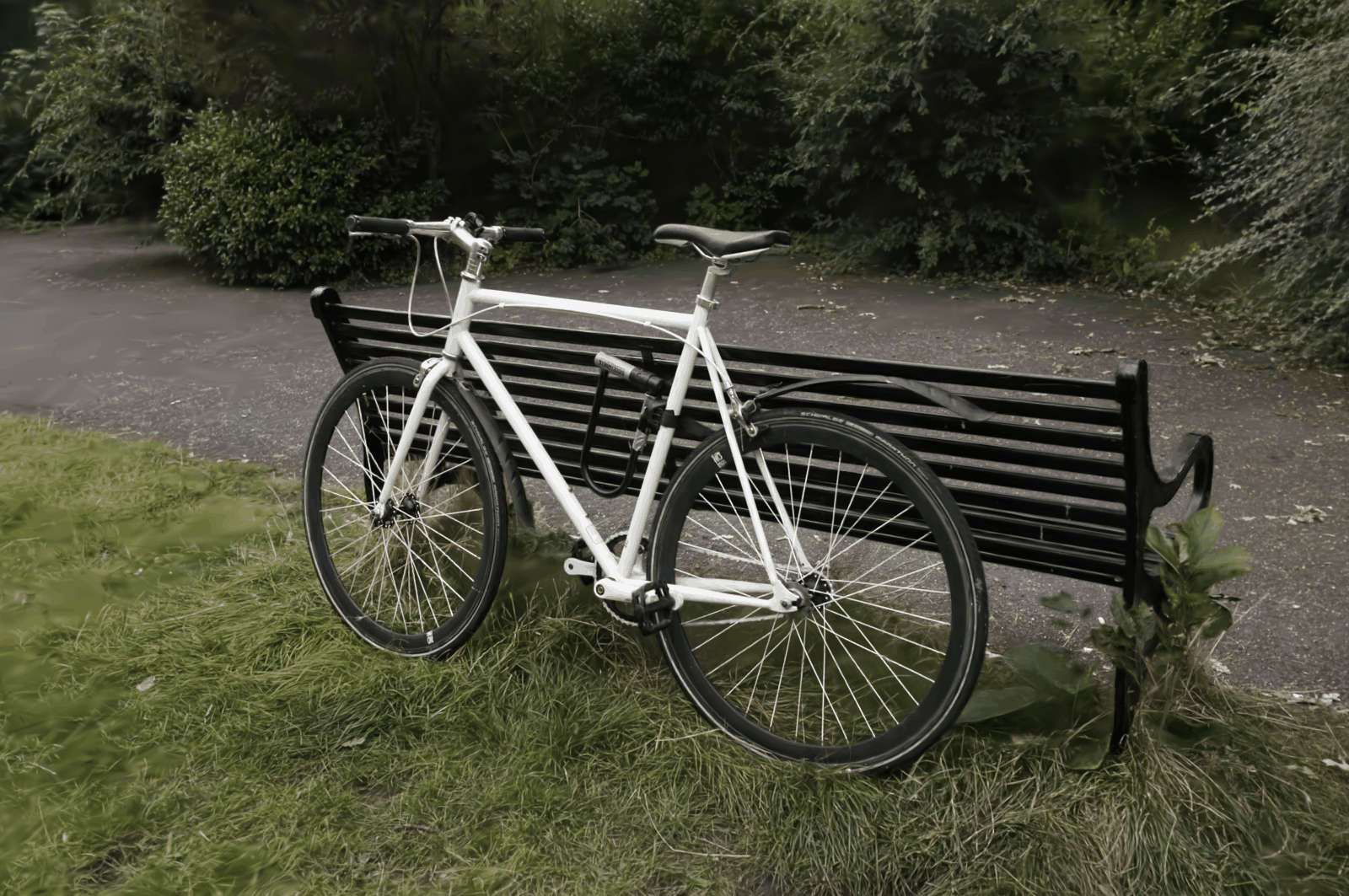}{3.0}{1.1}{0.55cm}{0.55cm}{1cm}{\mytmplen}{red}{26.06 / 24.70}&
        
        \\

		\zoomin{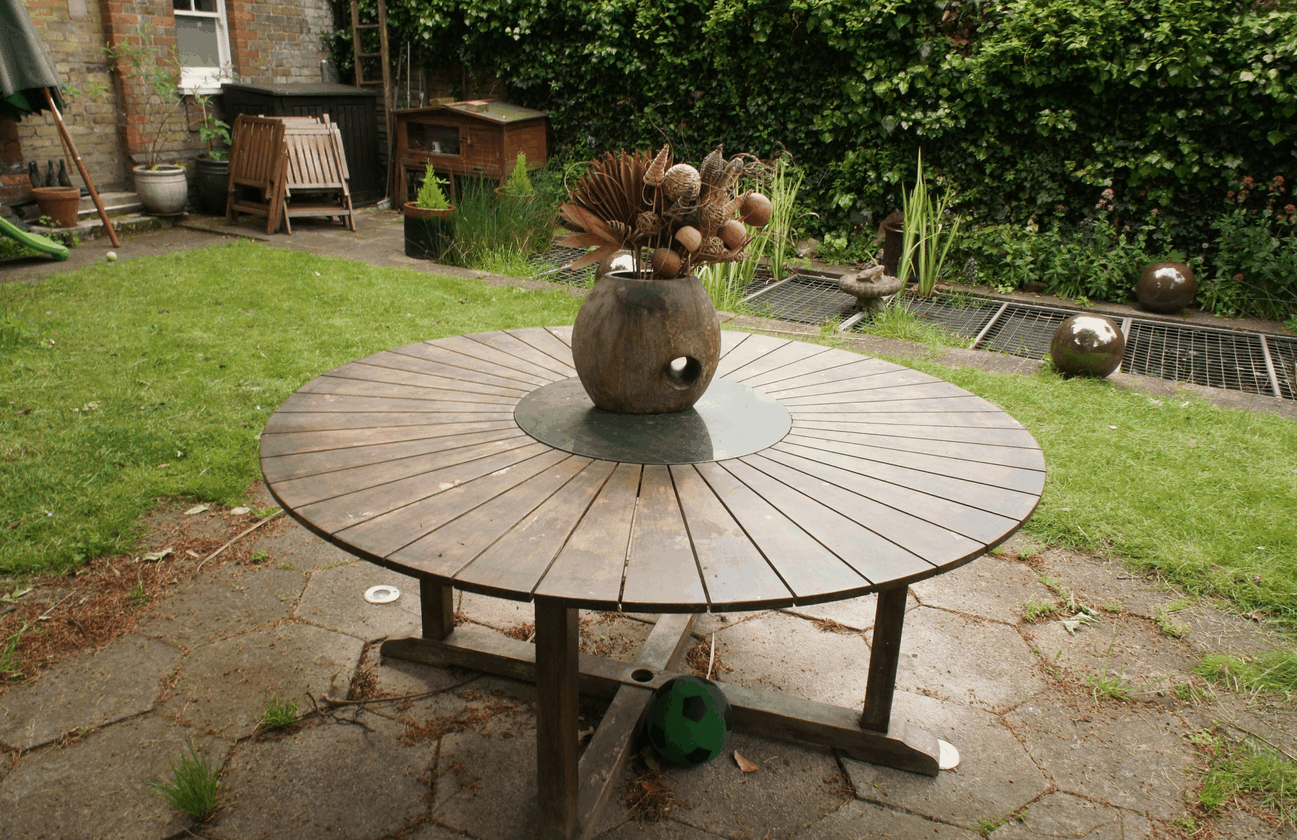}{0.6}{1.95}{0.55cm}{0.55cm}{1cm}{\mytmplen}{3}{red}&
		
        \zoominwithcomment{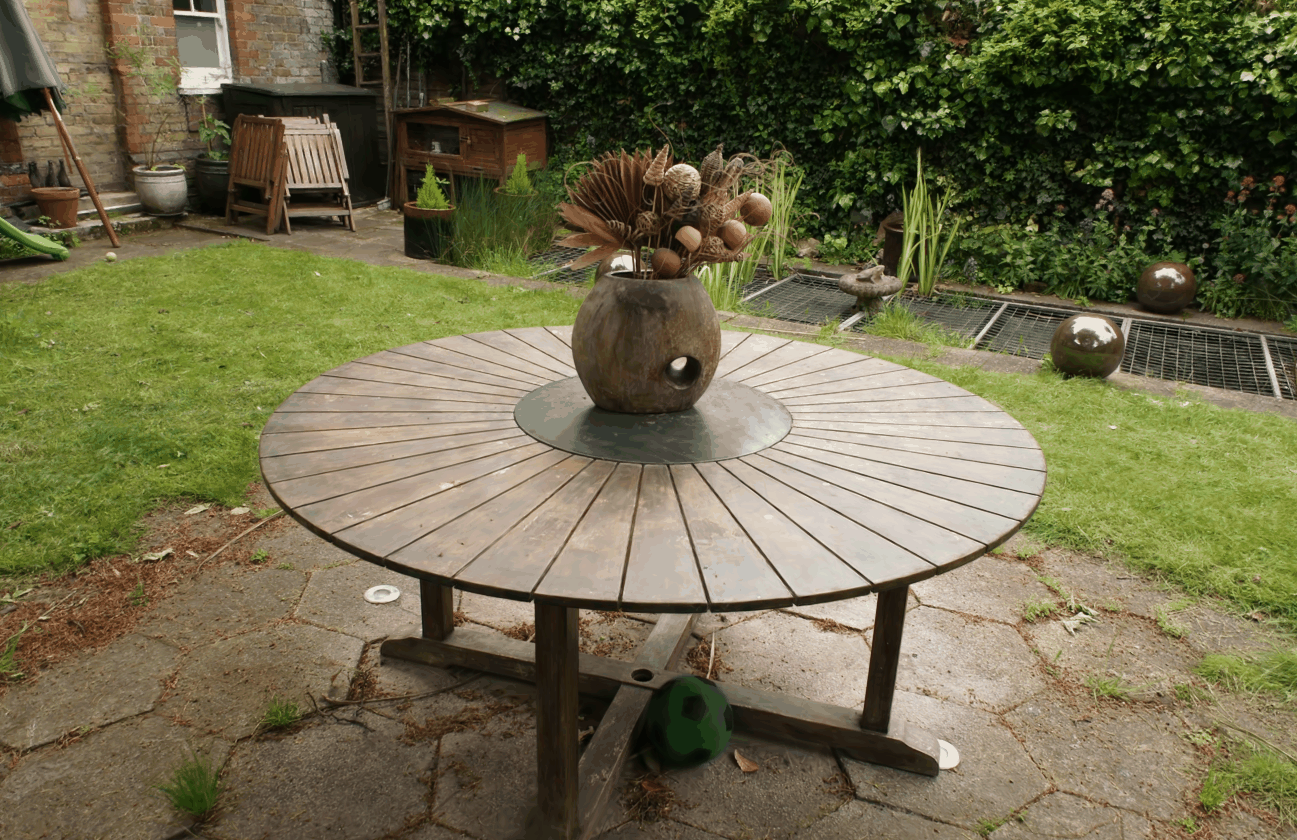}{0.6}{1.95}{0.55cm}{0.55cm}{1cm}{\mytmplen}{red}{27.56 / 26.29}&
		
        \zoominwithcomment{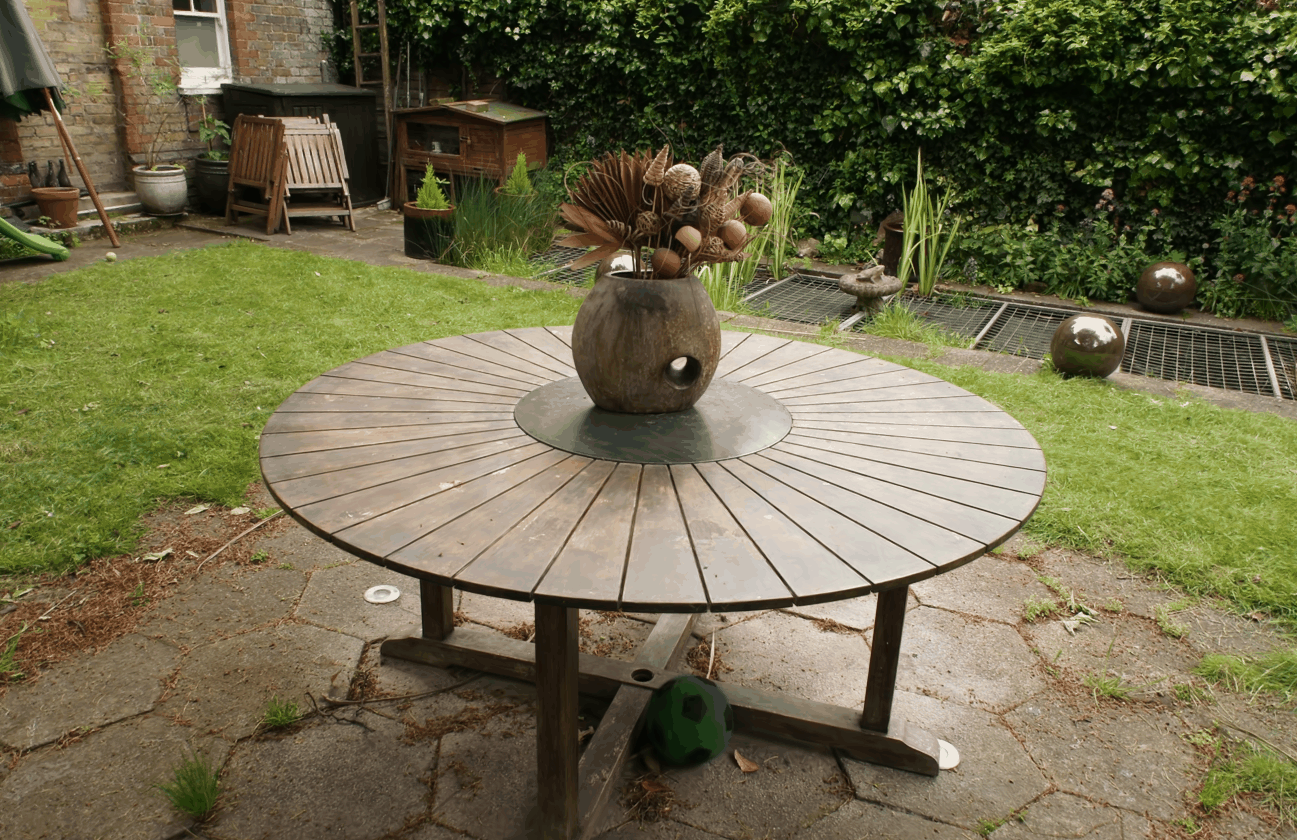}{0.6}{1.95}{0.55cm}{0.55cm}{1cm}{\mytmplen}{red}{29.06 / 27.18}&
		\zoominwithcomment{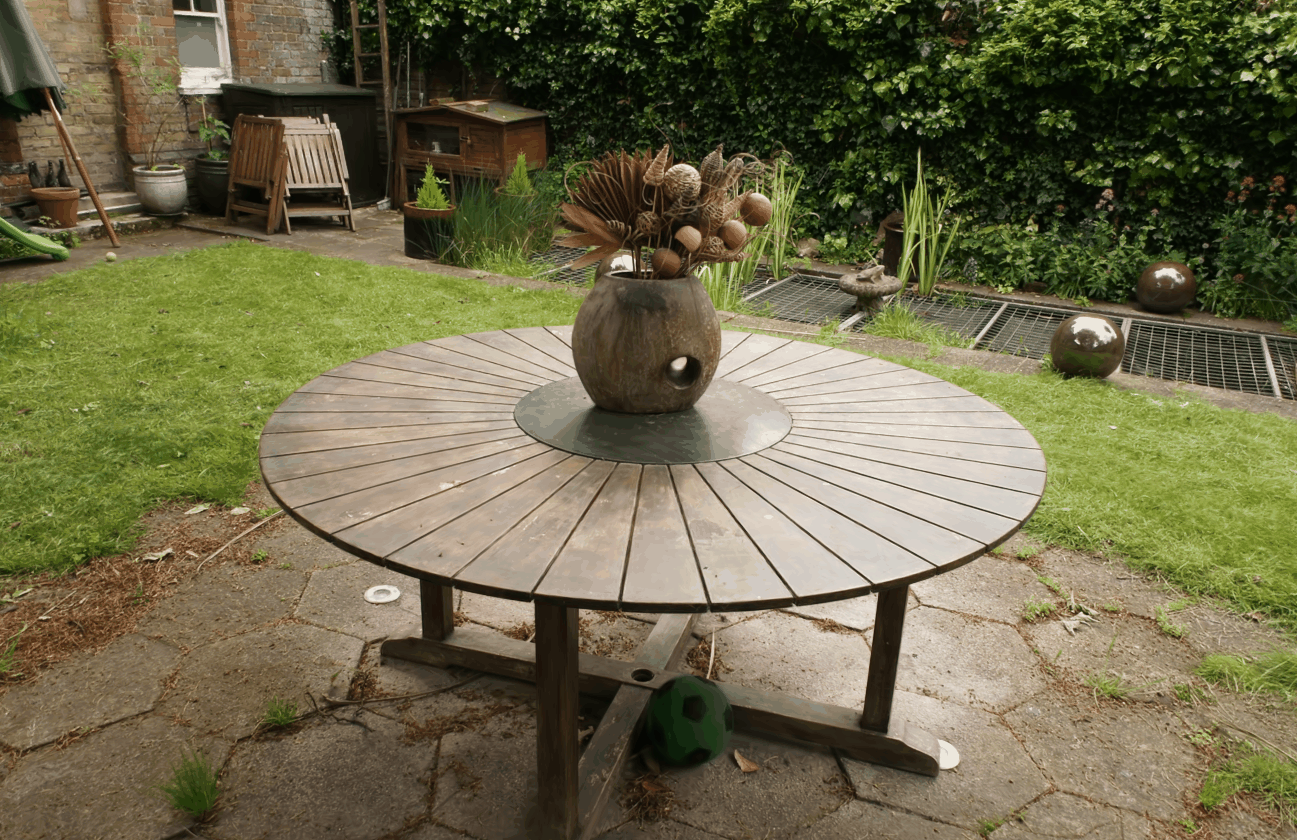}{0.6}{1.95}{0.55cm}{0.55cm}{1cm}{\mytmplen}{red}{28.69 / 26.75}&
		\zoominwithcomment{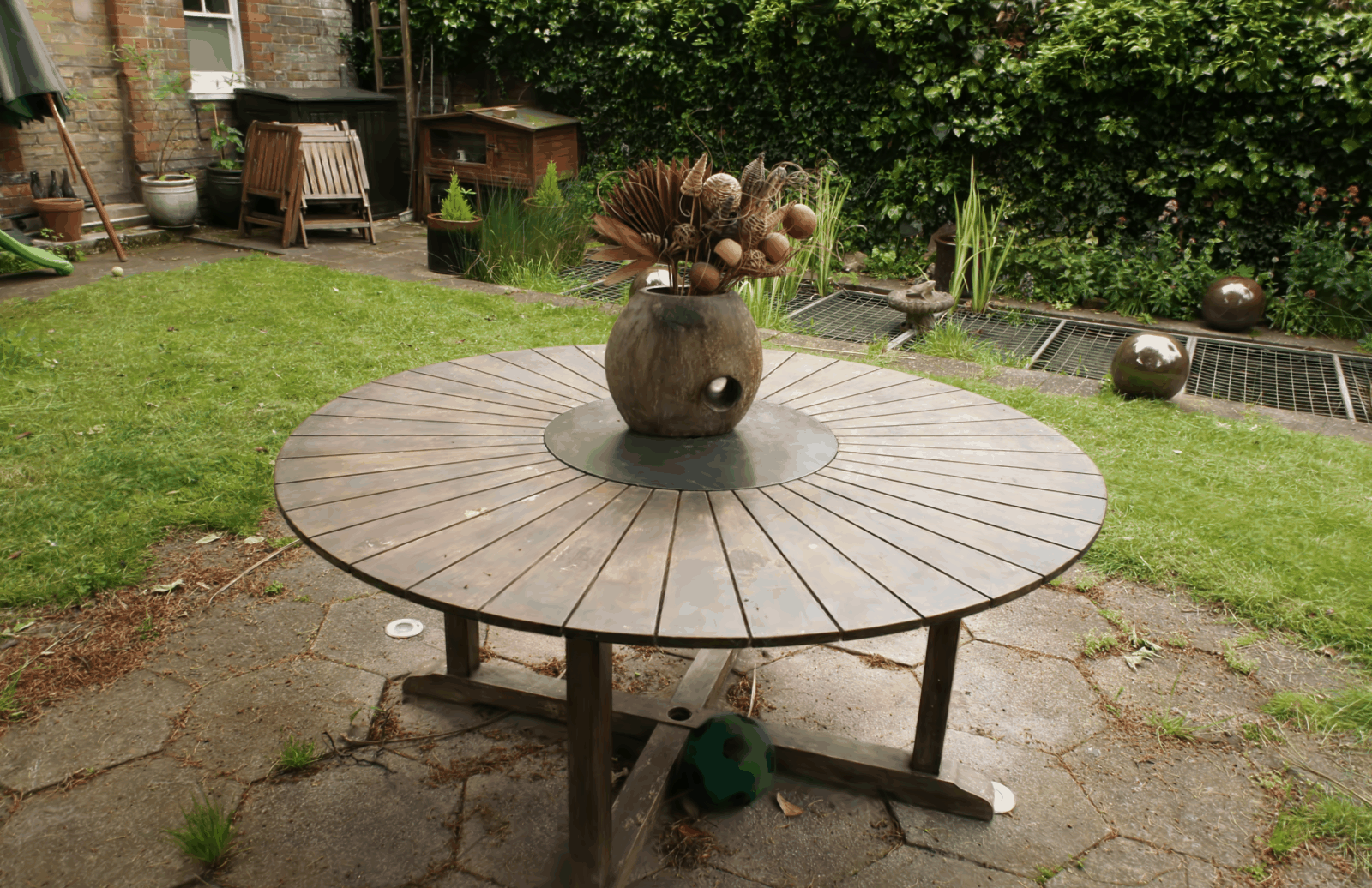}{0.6}{1.95}{0.55cm}{0.55cm}{1cm}{\mytmplen}{red}{28.17 / 27.45}
        
        \\
  
		\zoomin{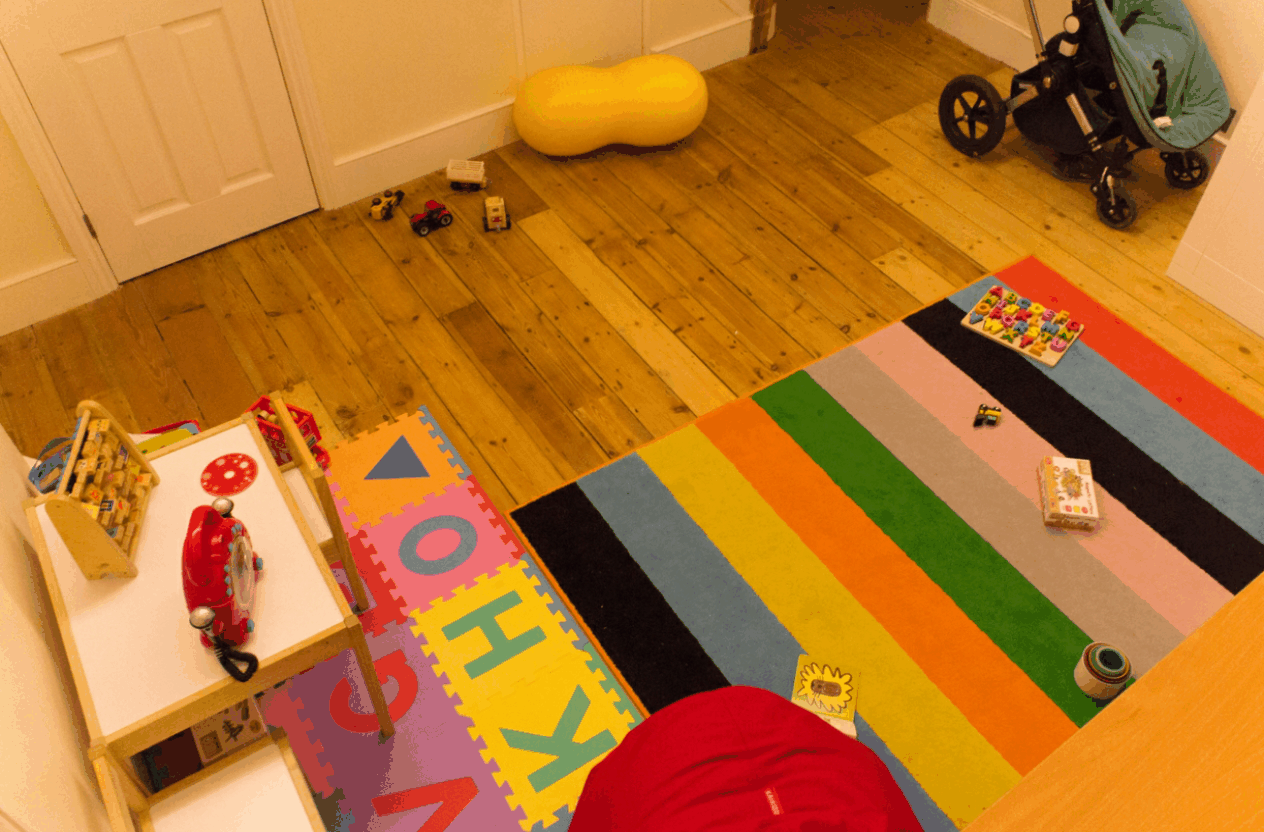}{1.2}{1.4}{0.55cm}{0.55cm}{1cm}{\mytmplen}{3}{green}&
		\zoominwithcomment{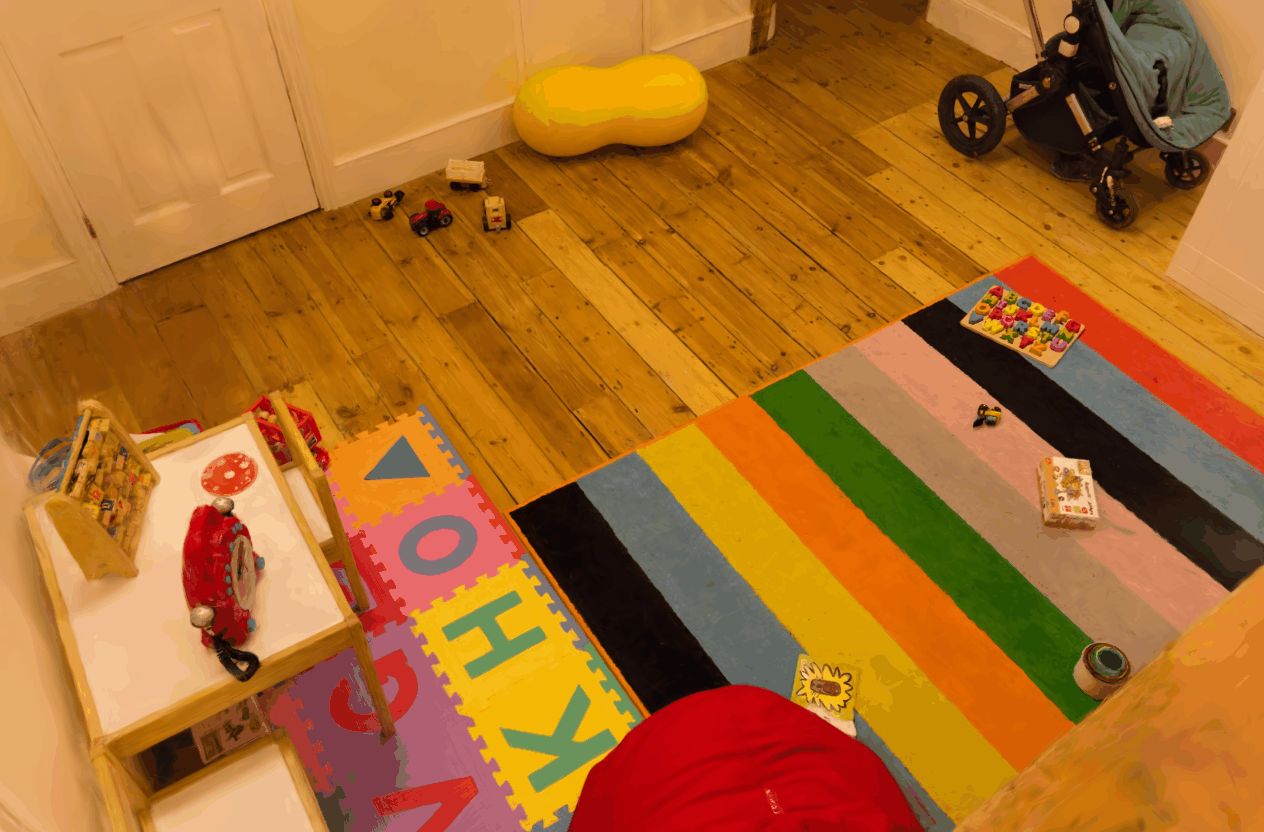}{1.2}{1.4}{0.55cm}{0.55cm}{1cm}{\mytmplen}{green}{25.62 / 29.11}&
		\zoominwithcomment{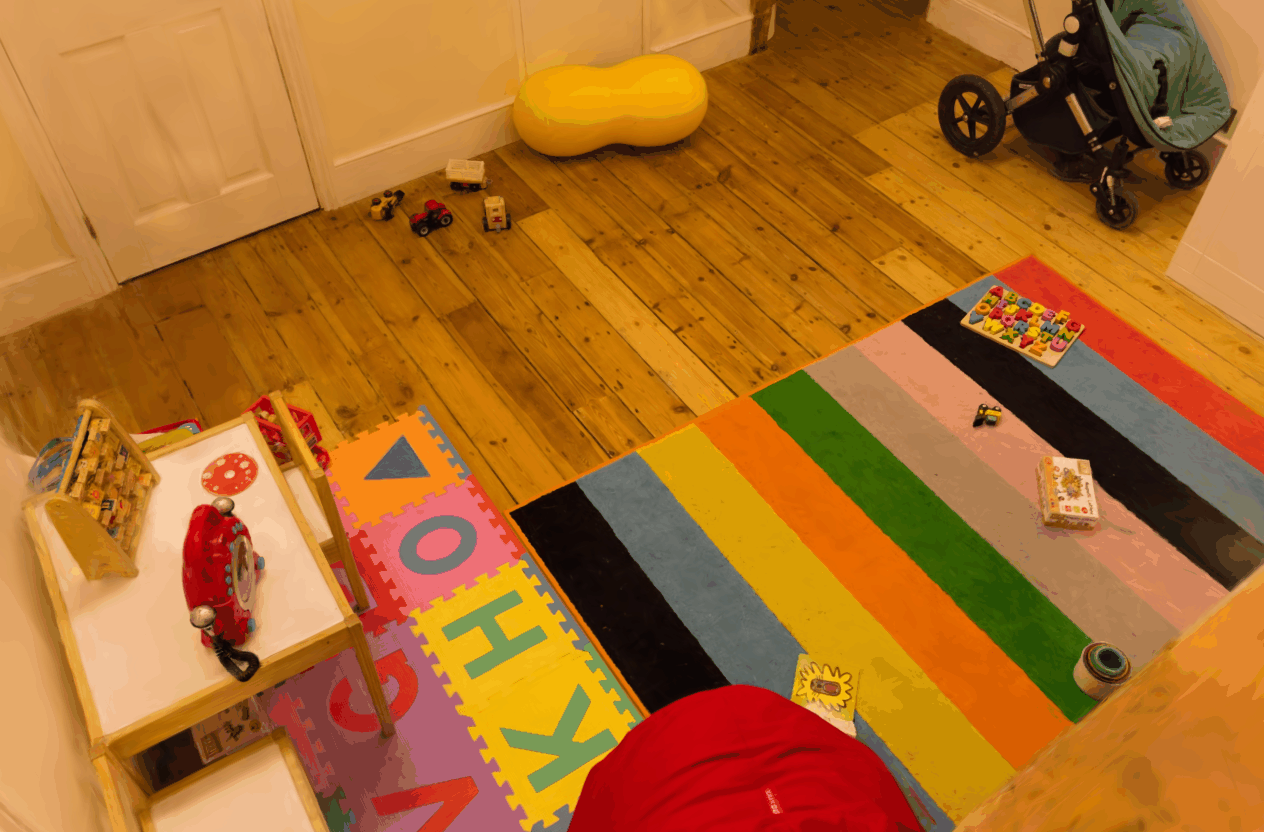}{1.2}{1.4}{0.55cm}{0.55cm}{1cm}{\mytmplen}{green}{25.72 / 29.93}&
		\zoominwithcomment{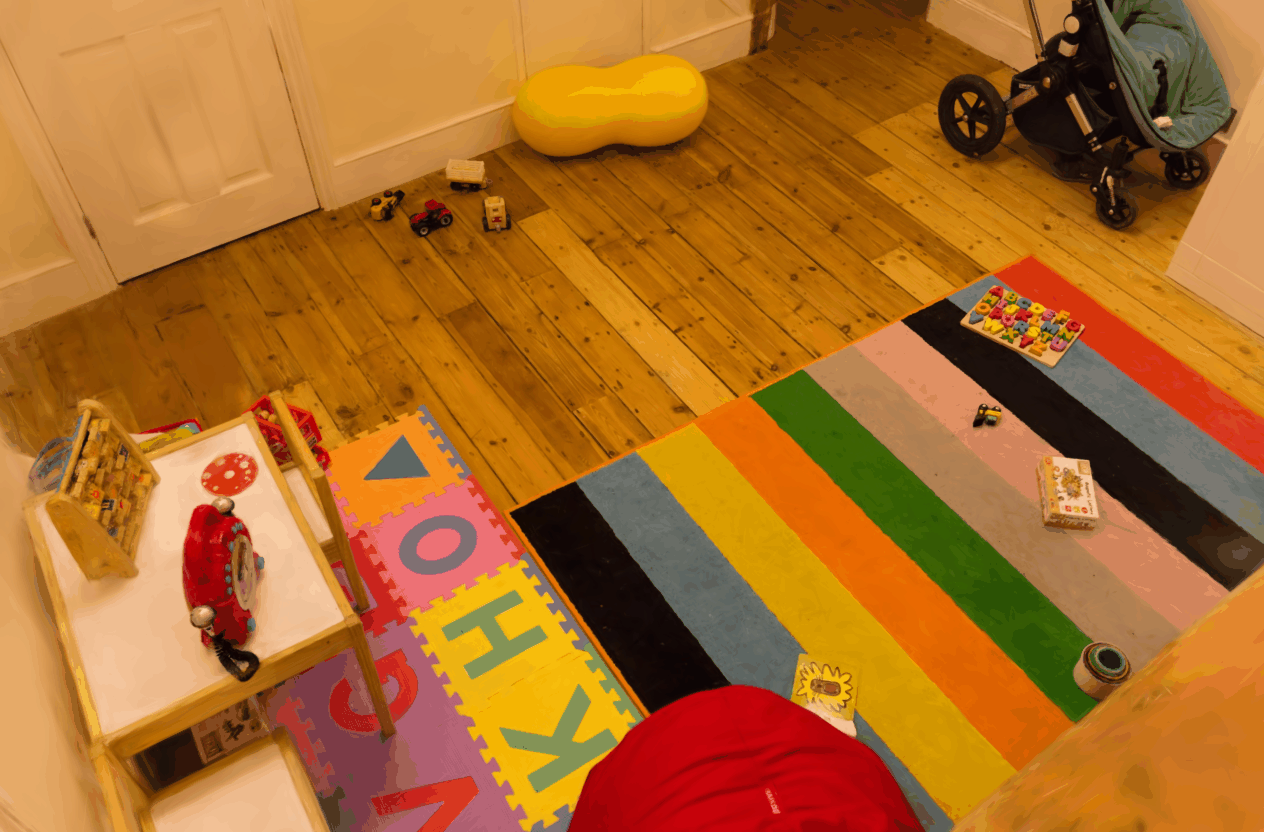}{1.2}{1.4}{0.55cm}{0.55cm}{1cm}{\mytmplen}{green}{25.85 / 29.89}& 
		\zoominwithcomment{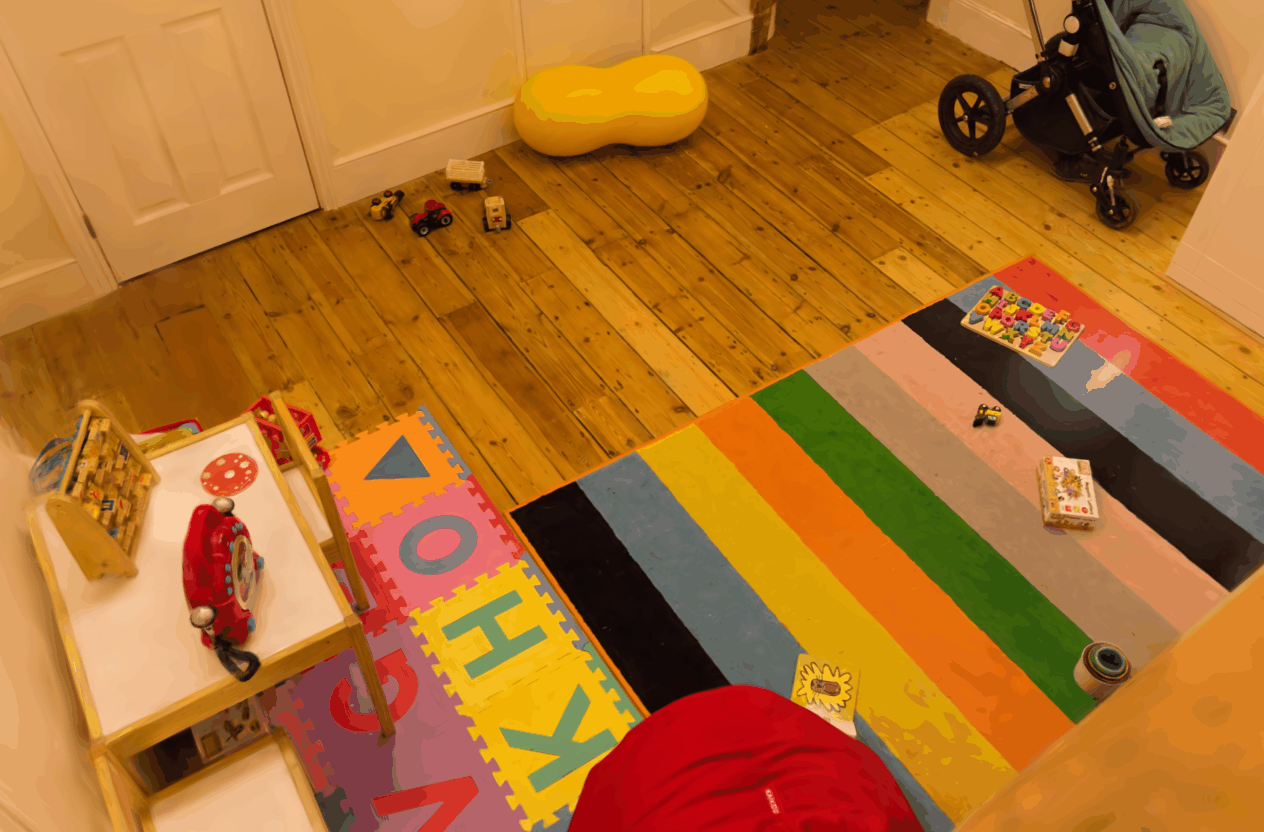}{1.2}{1.4}{0.55cm}{0.55cm}{1cm}{\mytmplen}{green}{25.74 / 30.04}
        
        \\

        \zoomin{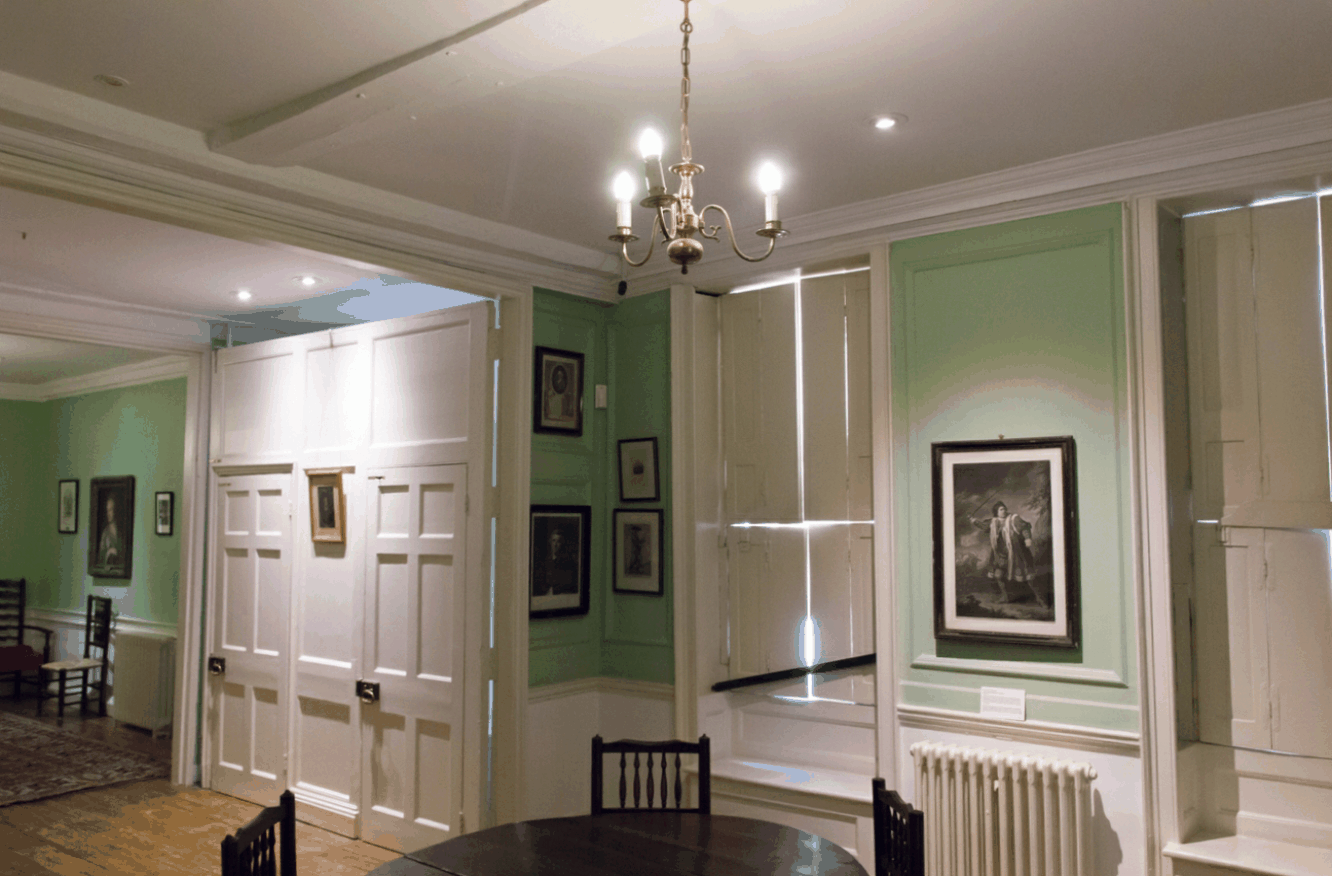}{2.7}{1.2}{0.55cm}{0.55cm}{1cm}{\mytmplen}{3}{green}&
		\zoominwithcomment{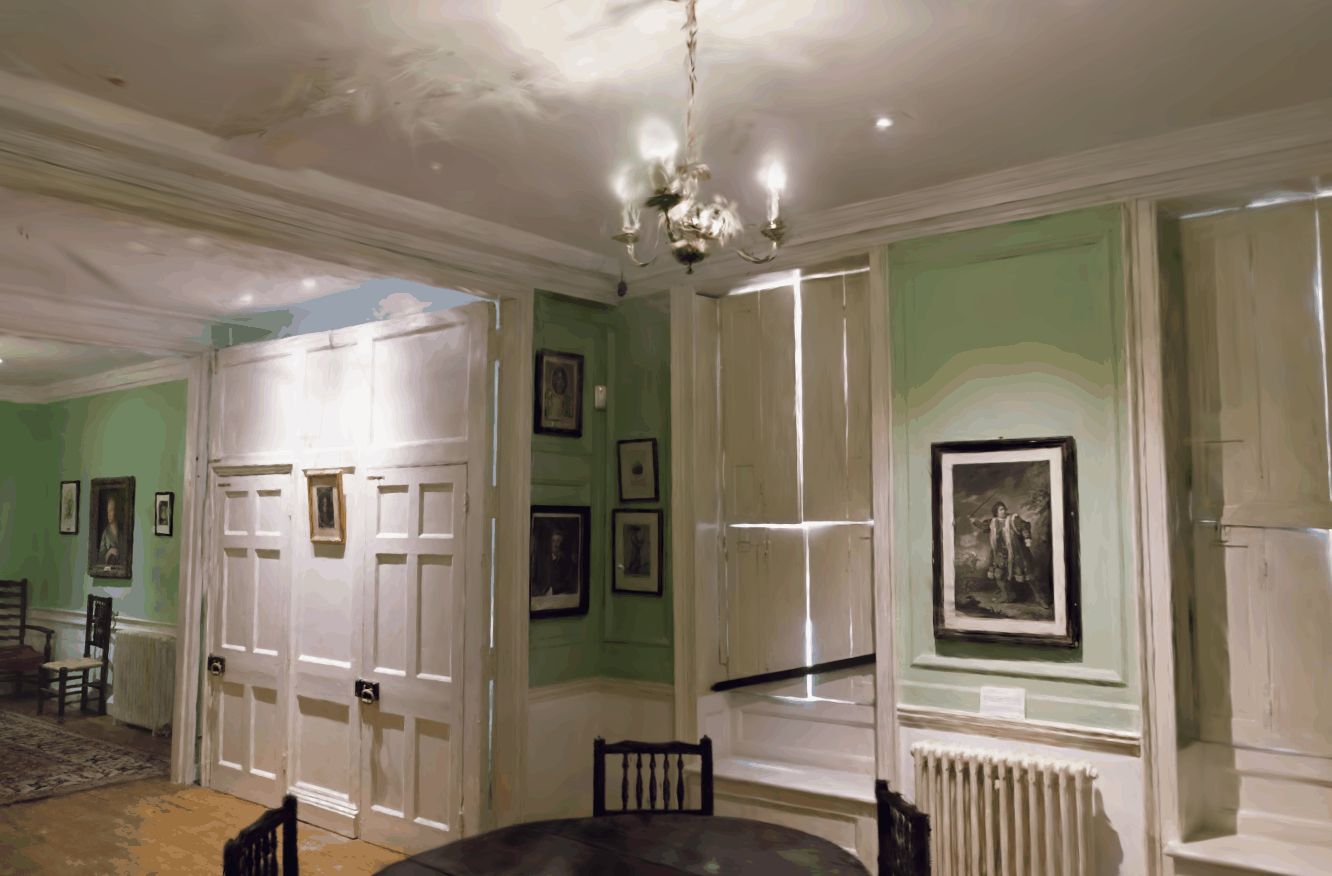}{2.7}{1.2}{0.55cm}{0.55cm}{1cm}{\mytmplen}{green}{25.53 / 28.11}&
		\zoominwithcomment{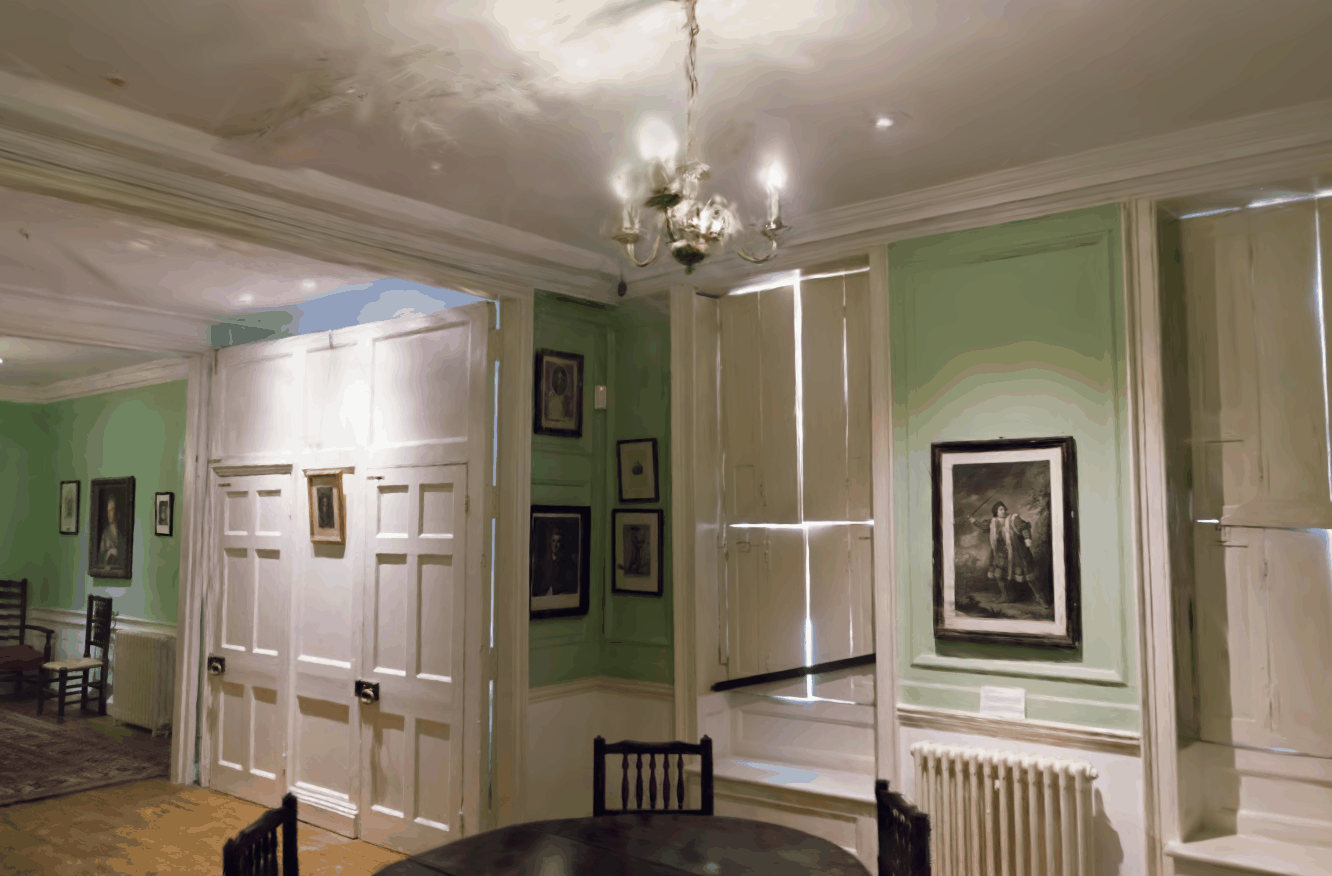}{2.7}{1.2}{0.55cm}{0.55cm}{1cm}{\mytmplen}{green}{25.76 / 28.94}&
		\zoominwithcomment{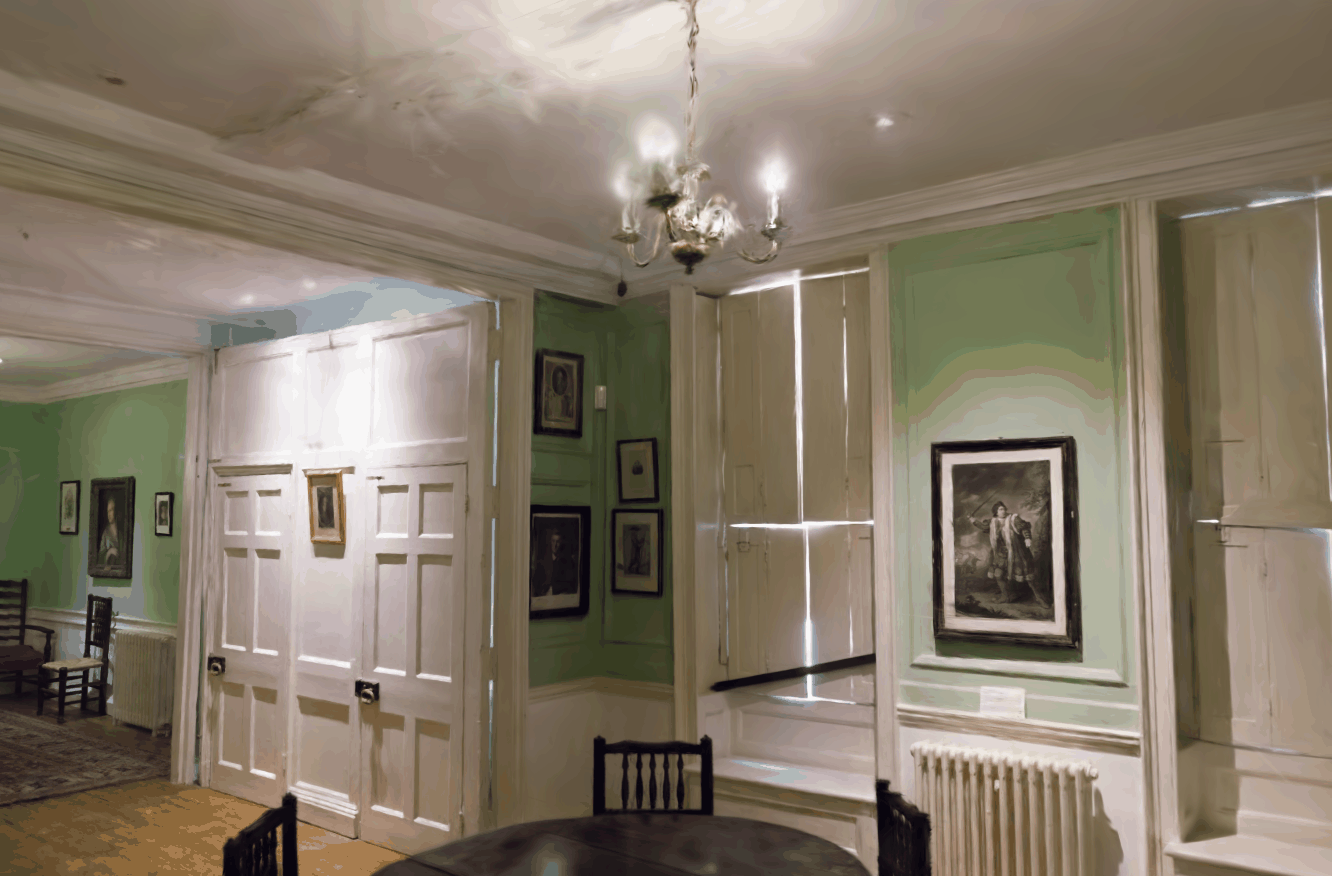}{2.7}{1.2}{0.55cm}{0.55cm}{1cm}{\mytmplen}{green}{25.70 / 28.87}& 
		\zoominwithcomment{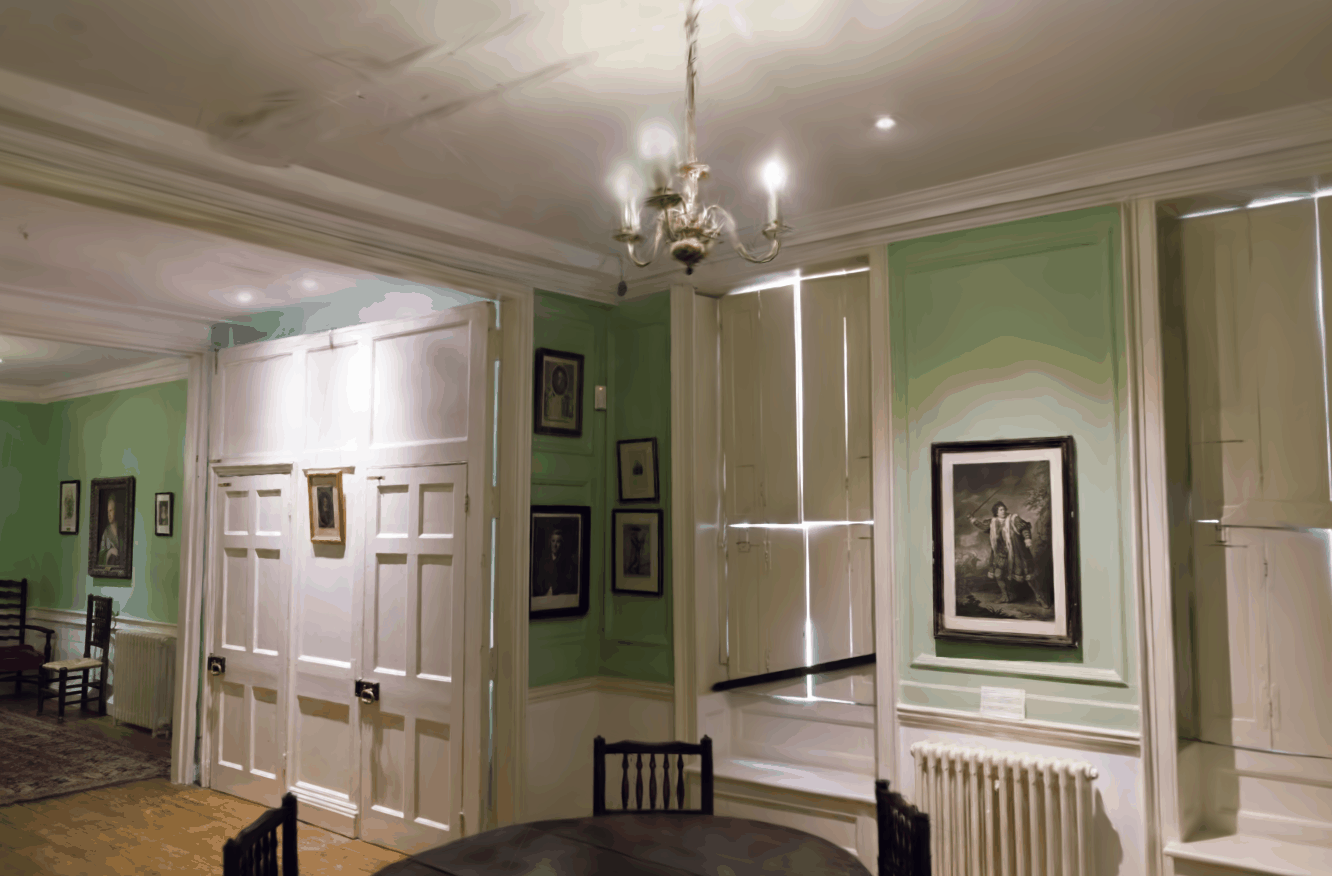}{2.7}{1.2}{0.55cm}{0.55cm}{1cm}{\mytmplen}{green}{26.41 / 29.33}
        
        \\

        \zoomin{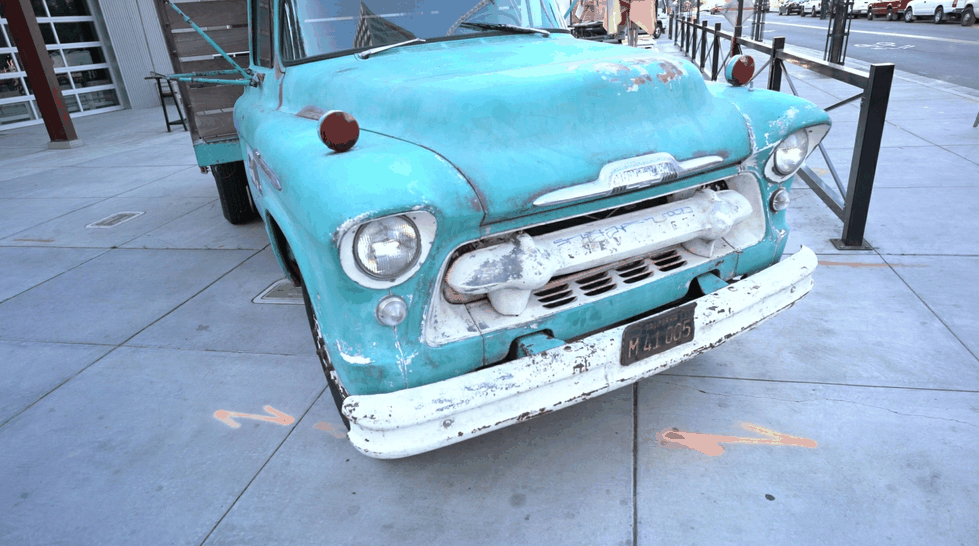}{2.7}{1.2}{0.55cm}{0.55cm}{1cm}{\mytmplen}{3}{blue}&
        
        \zoominwithcomment{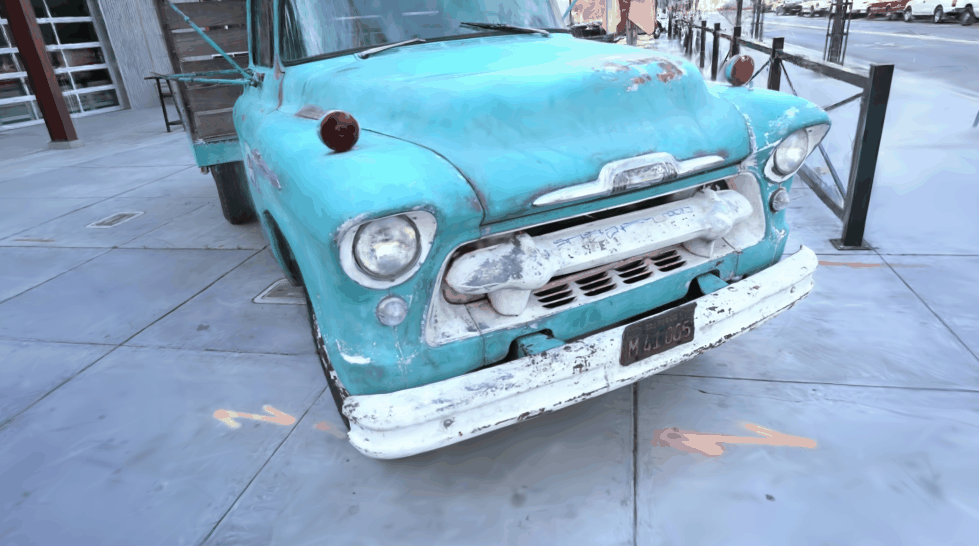}{2.7}{1.2}{0.55cm}{0.55cm}{1cm}{\mytmplen}{blue}{25.12 / 24.10}&
        
        \zoominwithcomment{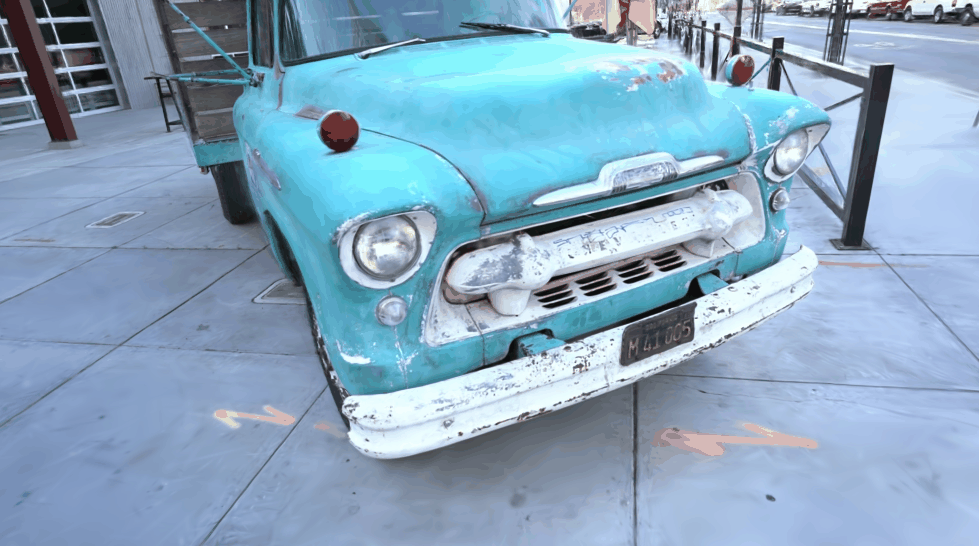}{2.7}{1.2}{0.55cm}{0.55cm}{1cm}{\mytmplen}{blue}{25.93 / 24.94}&
        \zoominwithcomment{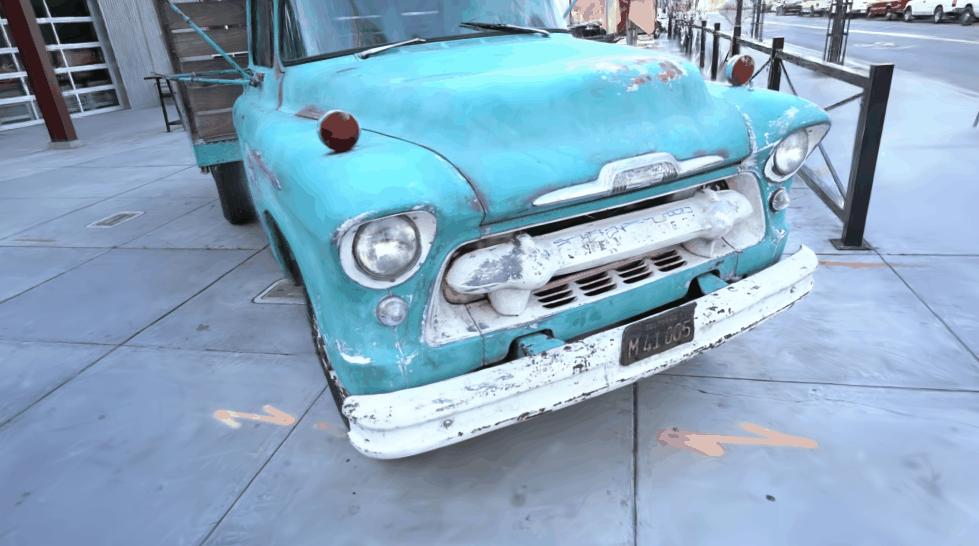}{2.7}{1.2}{0.55cm}{0.55cm}{1cm}{\mytmplen}{blue}{25.80 / 24.82}&
        
        \zoominwithcomment{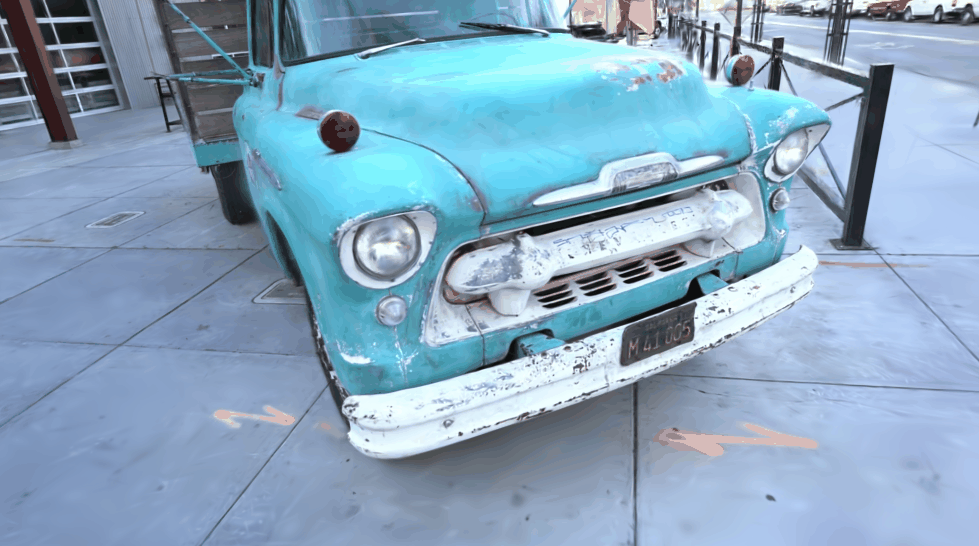}{2.7}{1.2}{0.55cm}{0.55cm}{1cm}{\mytmplen}{blue}{26.08 / 25.28}&

        \\
            
        \zoomin{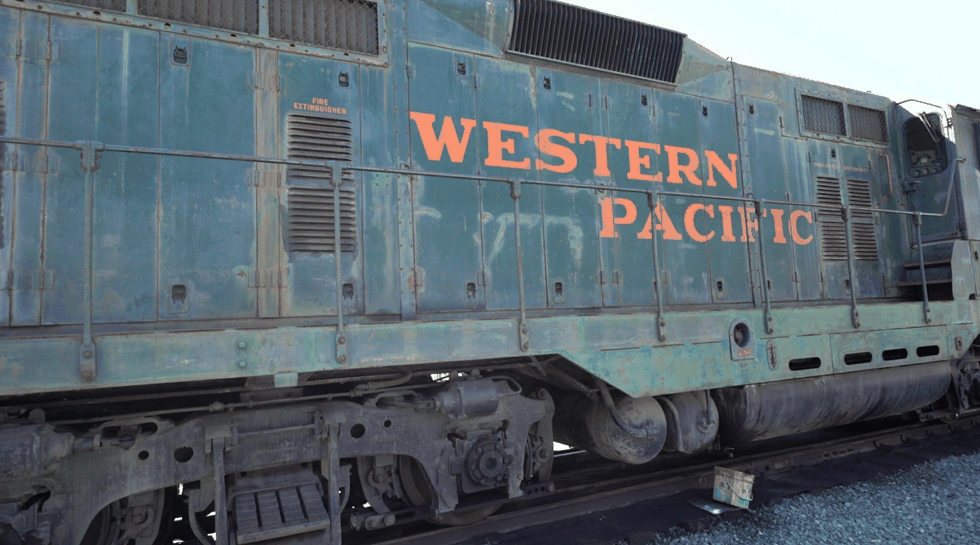}{2.1}{1.7}{0.55cm}{0.55cm}{1cm}{\mytmplen}{3}{blue}&
        
        \zoominwithcomment{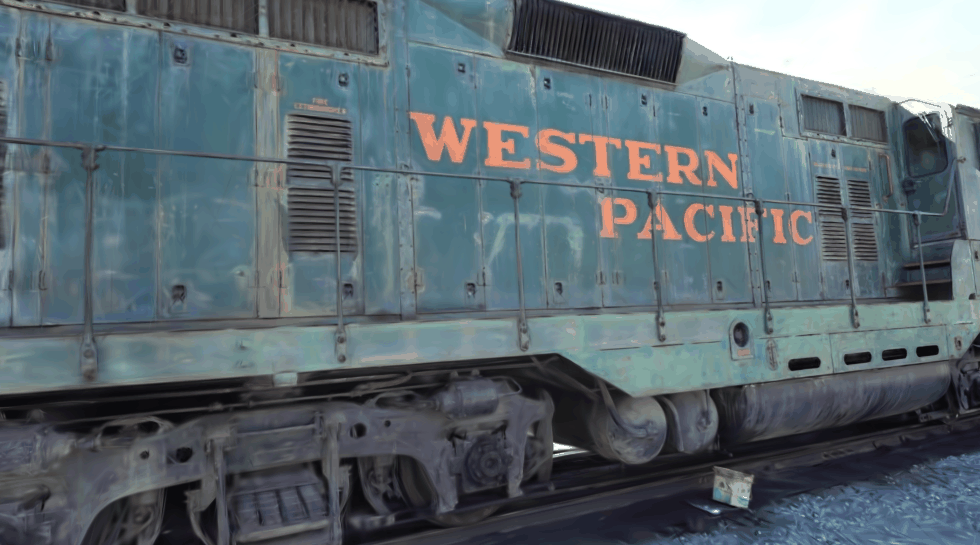}{2.1}{1.7}{0.55cm}{0.55cm}{1cm}{\mytmplen}{blue}{26.25 / 20.77}&
        
        \zoominwithcomment{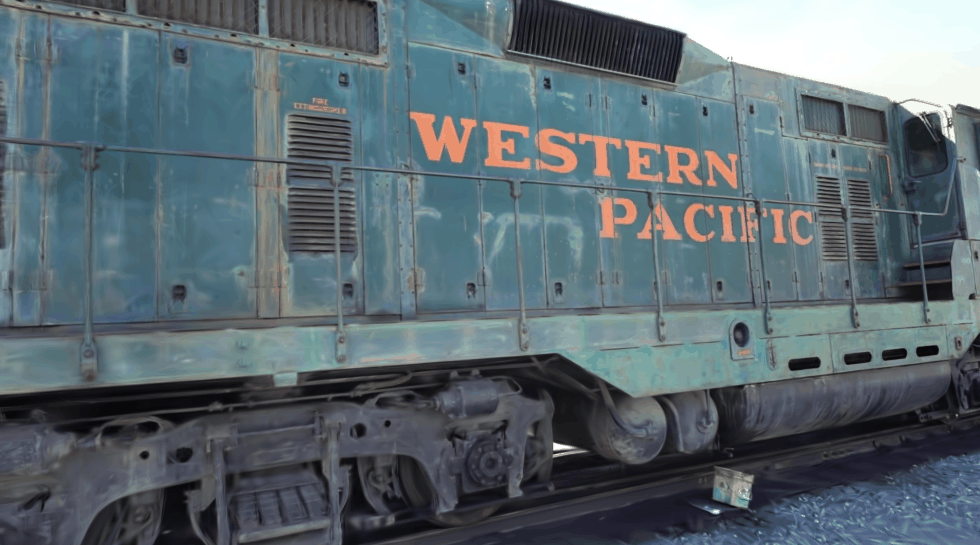}{2.1}{1.7}{0.55cm}{0.55cm}{1cm}{\mytmplen}{blue}{27.11 / 21.77}&
        \zoominwithcomment{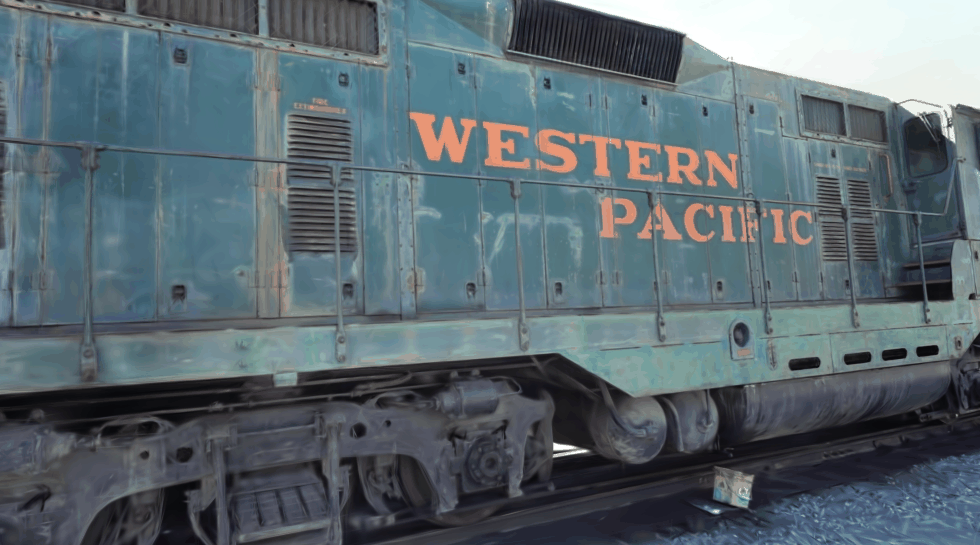}{2.1}{1.7}{0.55cm}{0.55cm}{1cm}{\mytmplen}{blue}{27.46 / 21.86}&
        
        \zoominwithcomment{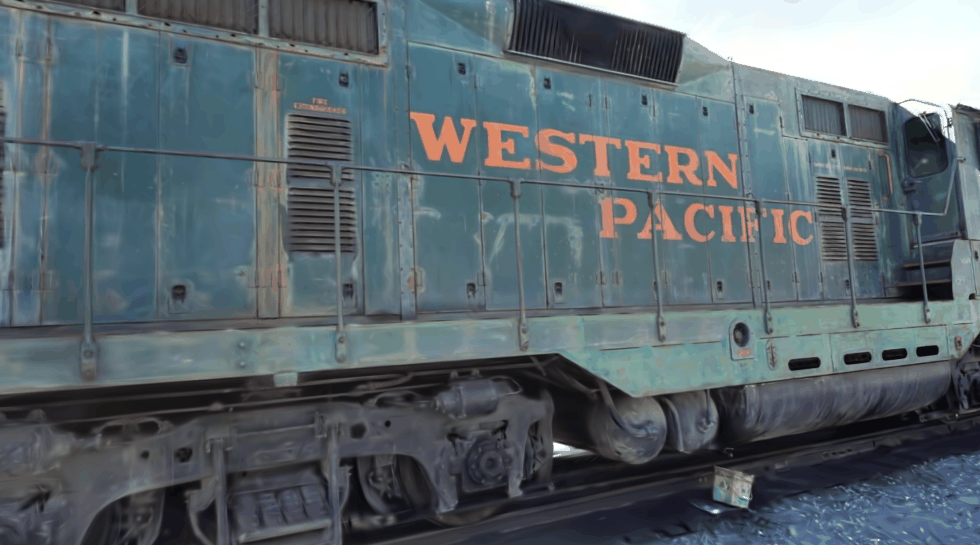}{2.1}{1.7}{0.55cm}{0.55cm}{1cm}{\mytmplen}{blue}{25.70 / 22.12}&
        
        \\
	\end{tabular}
	\caption{
		\label{fig:qualitative}
		We compare the visual quality of novel views synthesized from models in \name with the ground truth, \gaussian~\cite{kerbl20233d}, and two compression methods (refinement-based and retraining-based) using held-out test views. The evaluated scenes from top to bottom are \textsc{Bicycle} from Mip-NeRF360, \textsc{Playroom} from DeepBlending, and \textsc{Train} from Tanks\&Temples dataset. No obvious differences are observed between our method and the others. Results for other scenes are in the supplementary materials.
	}
\end{figure*}

\subsection{Main Results}

\noindent
\textbf{Quantitative results.}
We evaluate the quality, time cost, and output file size of all methods in \tref{tab:quant}. \name achieves compression quality and file sizes comparable to training-involved methods, with a time cost of 1\%-10\%, typically under a minute. A similar trend holds against the FCGS variant, FCGS-Opt, which eliminates training but is still 1.7-2.1$\times$ slower than our method. Specifically, \name reduces 94.9\%, 96.1\%, and 96.4\% of the data across three datasets, with quality losses of 0.8 dB, 0.7 dB, and 0.8 dB, respectively — all well below the 1 dB PSNR drop constraint, and with an average time of under 30 seconds. Results for \name on each scene are in \fixme{\tref{tab:per-scene-all}}, along with a detailed comparison to FCGS and its variants. These varying compression ratios and quality losses show that \name effectively identifies optimal compression parameters within the specified constraints. While training-involved methods offer a great balance of compression performance, they require pre-trained models and pose significant challenges, especially when training large-scale models.

Note that the time cost corresponds to the total execution time of the compression, including post-processing time for refinement or training time for retraining approaches. For our method, the time includes data I/O, importance score computation, and \foa. This comparison might seem unfair to retraining-based methods, as all others start with pre-trained models. However, we note that \name remains faster than all others, even when including the time required to train the \gaussian models (\gaussian + \name).

\textbf{Qualitative results.} \name achieves visually indistinguishable results when comparing images rendered from uncompressed (\gaussian~\cite{kerbl20233d}) and compressed models (\fref{fig:qualitative}). We evaluate scenes from all three datasets, including indoor, outdoor, intricate, and large objects, with PSNR scores overlaid for the selected view and averaged across the scene. Additionally, we provide rendered videos in the \fixme{supplementary materials} that support these findings.

\begin{figure}[ht!]
    \centering
    \includegraphics[width=\columnwidth]{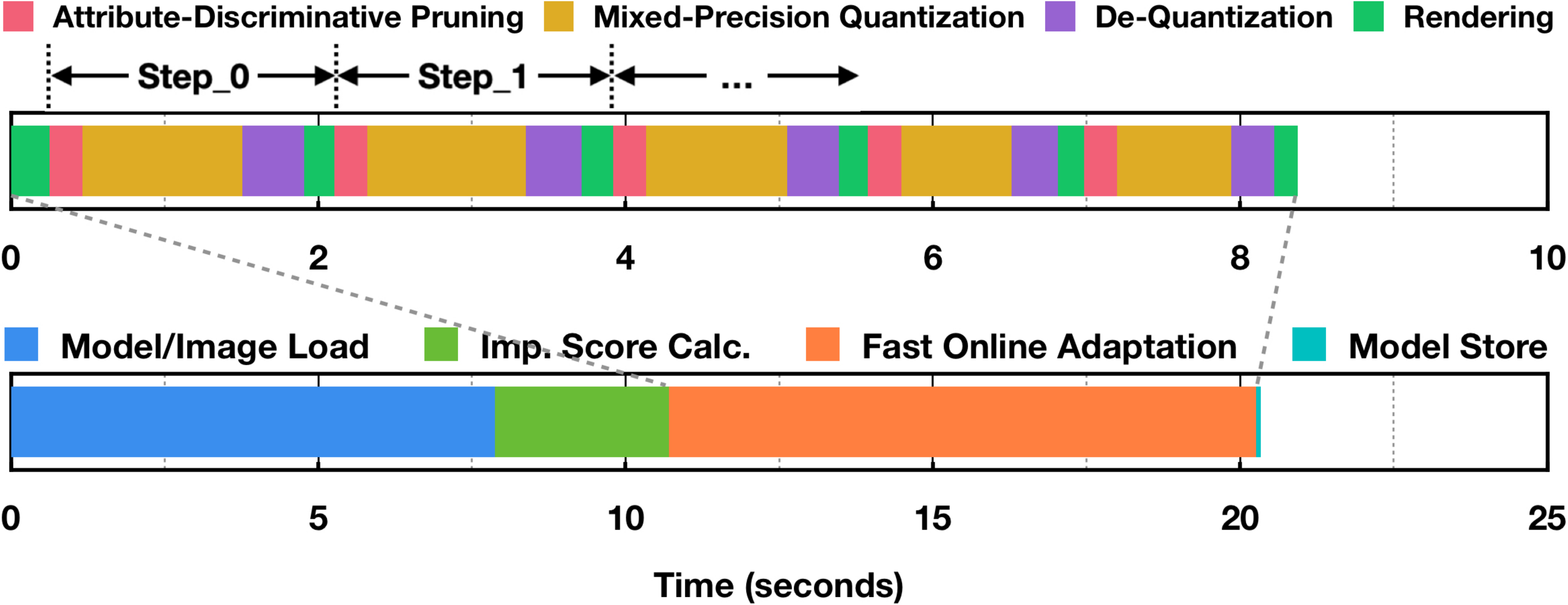}
    \caption{Fast online adaptation (\foa) adjusts quality and compression ratio every 2 seconds. The entire pipeline completes in under a minute, with data I/O being a non-trivial factor.}
    \label{fig:tuning}
\end{figure}

\textbf{Time breakdown.}
We break down the total compression time of \name, focusing on \foa’s online adaptation using the \textsc{Truck} scene in \fref{fig:tuning}. Time measurement begins with the one-time cost of loading pre-trained \gaussian models and camera parameters to compute the importance score. Only complete evaluation sets are needed for quality loss calculation, while training camera poses suffice for importance computation. A baseline rendering is performed at the start of \foa for quality reference. Each \foa step deep copies the input model at the millisecond level before irreversible pruning, followed by quantization, dequantization, and quality evaluation, which together take about one second. The process repeats with updated parameters from a limited search set, all starting with the input model, until the optimal compression setup that satisfies the specified constraints is found. The full compression takes about \fixme{20.3} seconds, with \fixme{38.8\%}, \fixme{13.9\%}, \fixme{47.1\%}, and \fixme{0.2\%} of the time spent on data loading, importance score calculation, online adaptation, and model storage, respectively. Each adaptation step takes about \fixme{1.56} seconds on average, with pruning, quantization, dequantization, and rendering taking \fixme{0.22}, \fixme{0.84}, \fixme{0.32}, and \fixme{0.18} seconds, respectively. The rendering time reflects rendering 32 views at approximately \fixme{178~FPS}. Fast adaptation to varying compression trade-offs is crucial for streaming Gaussians over fluctuating networks and on devices with diverse capabilities. Note that the times reported include the entire compression pipeline, with data I/O being a non-trivial part of our method but less significant for others.

\begin{figure}[t!]
  \begin{minipage}[t]{0.48\columnwidth}
    \centering
    \includegraphics[width=\columnwidth]{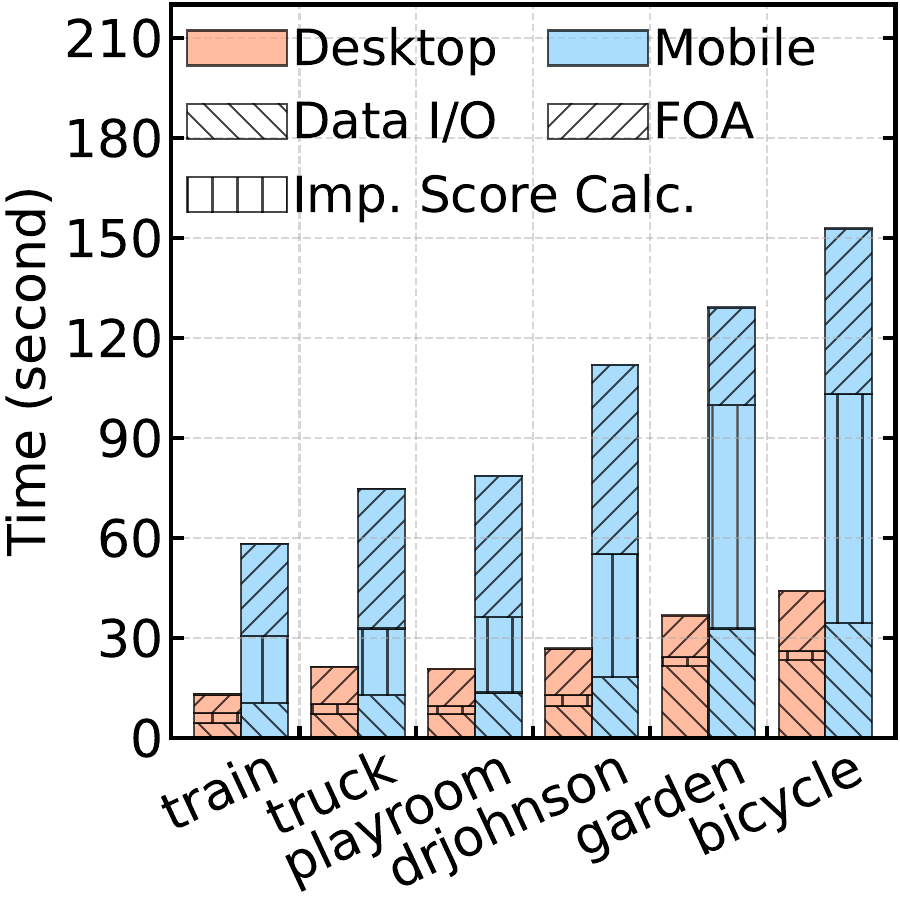}
    \caption{Time breakdown on desktop and mobile devices. Gaussian-intensive ops, like Imp. Score Calc., are more impacted by computing power.}
    \label{fig:cross-platform}
  \end{minipage}
  \hfill
  \begin{minipage}[t]{0.48\columnwidth}
    \centering
    \includegraphics[width=\columnwidth]{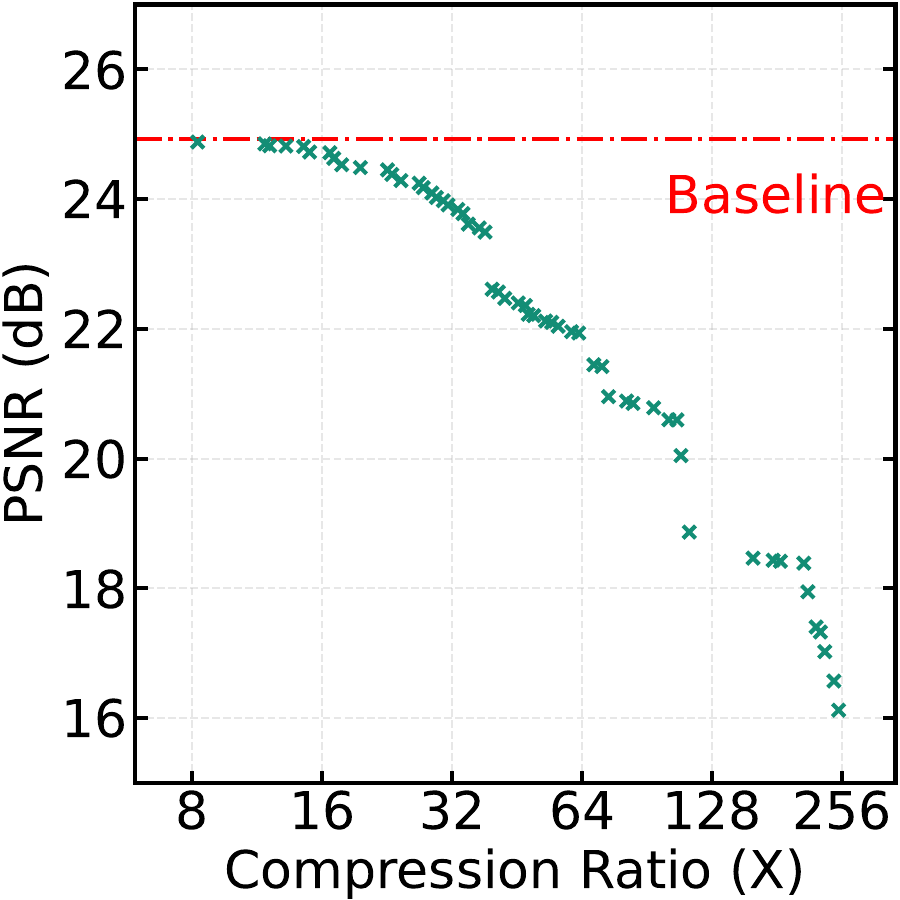}
    \caption{Rate-distortion performance of \name, allowing instant access to compressions from 8 to 256$\times$ and PSNR losses of 0.05 to 9 dB.}
    \label{fig:rate-distortion}
  \end{minipage}
\end{figure}

\textbf{Hardware costs.}
\name’s elimination of training also reduces hardware requirements, making it arguably the first compression method \textbf{deployable on mobile platforms}, where both FCGS~\cite{chen2024FCGS} (which runs out of 24 GiB VRAM even on a desktop setup, as shown by \oom{} in \fixme{\tref{tab:per-scene-all}}) and training-involved methods are impractical. \fref{fig:cross-platform} shows the time breakdown for a mobile setup with the Nvidia Jetson Xavier, compared to a desktop setup with an RTX 3090. With only 512 CUDA cores in the \textsc{Volta} architecture and 12 GB of shared memory (4 GB of 16 occupied by the OS and other processes), our training-free design completes compression in minutes, \fixme{detailed in \tref{tab:per-scene-all}}, still outperforming even the fastest Compressed3D~\cite{compressed-3dgs} on the desktop. The increased time is mainly due to data I/O and importance score computation ($\sim$7-24$\times$), rather than \foa ($\sim$3-5$\times$), highlighting that Gaussian-intensive processing, which is heavily involved in training, is more demanding on less powerful hardware. Additionally, the extra memory required for training data and optimizer states, which far exceed the available memory on mobile devices, poses out-of-memory challenges when deploying training-involved methods on mobile platforms.

\textbf{Rate-distortion performance.} 
\name enables fast exploration of the broad trade-off space between quality and compression ratio by adjusting pruning ratios for $Row-P$, $SH-P$, and bit-widths for attribute channels. While we set the quality constraint to $<$1 dB in our main experiments, the rate-distortion characteristics offer flexibility to optimize for either smaller file sizes or higher quality, spanning compression ratios from 8 to 256$\times$ and PSNR losses from 0.05 to 9 dB, as shown in \fref{fig:rate-distortion} for the \textsc{Truck} scene. Each set of configurable parameters corresponds to a unique design point in the trade-off space and can be reconfigured to any other point in about a second (i.e., a step in \fref{fig:tuning}). In contrast, most existing methods~\cite{fan2023lightgaussian, compressed-3dgs, navaneet2024compgs, pup-3dgs} are limited to fixed design points and struggle to adapt to varying compression needs, while methods like RDO-Gaussian~\cite{rdogaussian} and MesonGS~\cite{xie2025mesongs} require retraining or refinement to switch between design points. Despite being training-free, FCGS~\cite{chen2024FCGS} still requires a full feedforward pass (30-60 seconds) to reach a design point, while \foa in \name handles such adjustments, with each taking only 1-2 seconds.

\subsection{Analysis Results}

\noindent
\textbf{Large-scale \gaussian.}
\name eliminates the computation and time costs associated with other training-involved methods, offering greater efficiency and applicability, particularly when training \gaussian models requires specialized software or hardware. Large-scale Gaussian training with high-resolution data is increasingly used to represent larger scenes with high quality~\cite{turbo-gs, grendel-gs}, but it demands long training times, specialized software frameworks, and multi-GPU setups. We evaluate \name’s applicability by compressing large models trained with Grendel-GS~\cite{grendel-gs}. We train the \textsc{Bicycle} scene at various resolutions—from its original (around 4$K$) to downsampled versions (2$\times$, 4$\times$, and 8$\times$)—and at full resolution with checkpoints at different number of Gaussians, following Grendel-GS's densification approach. \tref{tab:large_scale} presents the time cost, peak memory usage, compression quality, and compression ratios for both training and applying \name, all measured on a single Nvidia A100 GPU, with the quality drop restricted to $<$1 dB.

\begin{table}[ht!]
    \caption{Training and compression costs vary with Gaussian scales, all measured under a quality constraint of $<$1 dB loss. High-resolution data and large Gaussians incur significant overhead in training and training-involved compression, evidencing the advantage of our training-free design.}
    \centering
    \huge
    \resizebox{\columnwidth}{!}{
    \begin{tabular}{l|cccc}
        \toprule[0.15em]
        ~ & \multicolumn{4}{c}{Training Resolution} \\
        ~ & 618$\times$411 & 1237$\times$822 & 2473$\times$1643 & 4946$\times$3286 \\
        \midrule[0.05em]

        No. Gaussians~(million) & 3.04 & 6.16 & 9.36 & 11.84 \\
        \midrule[0.05em]

        Training Time~(\textcolor{red}{min.}) & 17 & 33 & 73 & 190 \\ 
        Training Mem.~(GiB) & 6.81 & 14.00 & 23.54 & 39.37 \\
        PSNR (dB) & 26.54 & 24.70 & 24.49 & 24.44 \\
        \midrule[0.05em]
        \name Time~(\textcolor{red}{sec.}) & 20.08 & 38.64 & 64.71 & 104.31 \\ 
        \name Mem.~(GiB) & 3.89 & 6.75 & 12.39 & 
        21.04 \\ 
        PSNR (dB) & 25.64 & 24.10 & 23.72 & 23.89  \\
        Compression Ratio~($\times$) & 17.13 & 20.38 & 28.04 & 28.00 \\ 
        \bottomrule[0.15em]
    \end{tabular}
    }
    \label{tab:large_scale}
\end{table}

As image resolution and Gaussian counts increase, training costs in time and hardware rise sharply. This also challenges training-involved compression methods, which often require similar software and hardware resources. In contrast, \name compresses Gaussian models at various scales with minimal resource and time consumption, all within the 1 dB quality constraint. This is made possible by eliminating the training process and the need for additional memory to store training data and optimizer states.

\newcommand{\specialcell}[2][c]{\begin{tabular}[#1]{@{}c@{}}#2\end{tabular}}

\begin{table*}[h]
    \caption{\label{tab:per-scene-all} PSNR (dB), file size (MiB), and compression time (sec.) for each test scene. \oom{}: out of memory on RTX 3090 (24 GiB).}
    \huge
    \centering
    \resizebox{\textwidth}{!}{
        \begin{tabular}{ c|c|ccccccccc|cc|cc } 
            \toprule[0.20em]
            \multirow{2}{*}{Metrics} & \multirow{2}{*}{Method} & \multicolumn{9}{c|}{\textbf{Mip-NeRF360}} & \multicolumn{2}{c|}{\textbf{Tanks\&Temples}} & \multicolumn{2}{c}{\textbf{Deep Blending}} \\ 
             &  & Bicycle & Bonsai & Counter & Flowers & Garden & Kitchen & Room & Stump & Treehill & Train & Truck & Drjohnson & Playroom \\ 
            \midrule[0.05em]
            \multirow{5}{*}{\specialcell{PSNR\\(dB)}} & \gaussian~\cite{kerbl20233d} &
            25.18 & 32.02 & 28.91 & 21.46 & 27.17 & 30.81 & 31.40 & 26.6 & 22.29 & 21.78 & 24.93 & 28.99 & 30.13 \\
            & FCGS-Raw $1e^{-4}$ &
            \oom{} & 32.12 & 28.98 & \oom{} & \oom{} & 30.83 & 31.54 & \oom{} & \oom{} & 21.76 & \oom{} & \oom{} & 30.24 \\
            & FCGS-Opt $1e^{-4}$ &
            24.40 & 32.12 & 28.98 & 21.17 & 26.27 & 30.83 & 31.54 & 25.98 & 22.04 & 21.76 & 24.96 & 28.99 & 30.24 \\
            & FCGS-Opt $16e^{-4}$ &
            \oom{} & 31.49 & 28.67 & \oom{} & \oom{} & 30.08 & 31.07 & \oom{} & 22.04 & 21.65 & \oom{} & \oom{} & 29.98 \\
            & \cellcolor{pink!20}\name &
            24.18 & 31.05 & 28.24 & 20.57 & 26.30 & 30.03 & 30.67 & 25.72 & 21.29 & 20.83 & 24.16 & 28.13 & 29.29 \\
            \midrule[0.05em]
            \multirow{5}{*}{\specialcell{Size\\(MiB)}} & \gaussian~\cite{kerbl20233d} &
            1450.28 & 294.42 & 289.24 & 860.06 & 1379.99 & 438.10 & 376.85 & 1173.52 & 894.90 & 242.78 & 601.03 & 805.36 & 602.19 \\
            & FCGS-Raw $1e^{-4}$ &
            \oom{} & 24.32 & 24.29 & \oom{} & \oom{} & 39.17 & 26.48 & \oom{} & \oom{} & 20.12 & \oom{} & \oom{} & 49.10 \\
            & FCGS-Opt $1e^{-4}$ &
            92.60 & 24.31 & 24.29 & 61.72 & 113.23 & 39.17 & 26.48 & 101.70 & 77.56 & 20.12 & 42.12 & 61.71 & 49.10 \\
            & FCGS-Opt $16e^{-4}$ &
            \oom{} & 13.50 & 13.58 & \oom{} & \oom{} & 20.67 & 15.21 & \oom{} & 42.22 & 10.86 & \oom{} & 34.76 & 27.46 \\
            & \cellcolor{pink!20}\name &
            70.97 & 24.56 & 19.82 & 31.20 & 89.80 & 36.70 & 22.03 & 57.99 & 23.98 & 10.73 & 21.91 & 29.19 & 22.02 \\
            \midrule[0.05em]

            \multirow{5}{*}{\specialcell{Time\\(Second)}} & FCGS-Raw $1e^{-4}$ Desktop&
            \oom{} & 20.75 & 20.81 & \oom{} & \oom{} & 29.73 & 24.41 & \oom{} & \oom{} & 19.83 & \oom{} & \oom{} & 37.34 \\
            & FCGS-Opt $1e^{-4}$ Desktop&
            92.60 & 20.30 & 20.29 & 61.72 & 86.83 & 28.39 & 23.85 & 84.14 & 64.04 & 18.69 & 41.48 & 48.22 & 36.99 \\
            & FCGS-Opt $16e^{-4}$ Desktop&
            \oom{} & 13.15 & 12.94 & \oom{} & \oom{} & 18.65 & 16.36 & \oom{} & 40.60 & 11.20 & \oom{} & 33.72 & 25.18 \\
            & \cellcolor{pink!20}\name Desktop &
            37.75 & 21.56 & 15.27 & 27.42 & 30.25 & 25.66 & 15.74 & 28.60 & 29.00 & 13.52 & 22.96 & 28.93 & 22.36 \\
            & \cellcolor{pink!20}\name Mobile &
            135.49 & 94.37 & 68.34 & 91.33 & 117.18 & 122.48 & 82.22 & 66.31 & 82.21 & 59.42 & 73.88 & 113.45 & 77.15 \\
            \bottomrule[0.20em]
        \end{tabular}
    }
\end{table*}
\textbf{Comparison to FCGS.}
FCGS~\cite{chen2024FCGS} is a newly introduced Gaussian compression method that operates without refinement or retraining, making it highly relevant to our work. However, the off-the-shelf version fails on 7 out of 13 test scenes, requiring over 24 GiB of GPU VRAM—typical for GPUs like the RTX 3090 and A6000, as shown by the \oom{} error in the FCGS-Raw in \fixme{\tref{tab:per-scene-all}}, an inherent limitation of its design. To ensure a fair comparison across compression methods on the same hardware, we evaluate our enhanced variant, FCGS-Opt, with customized data partition strategies. FCGS-Opt reduces memory usage and execution time while maintaining the original compression quality and ratios. We follow FCGS's recommended setup, testing with two hyperparameters: $1e^{-4}$, prioritizing quality retention, and $16e^{-4}$, prioritizing smaller file sizes.

\fixme{\tref{tab:per-scene-all}} confirms that, under the same $1e^{-4}$ setting, FCGS-Opt reduces memory and time costs of FCGS-Raw at no compromise. However, it still fails due to out-of-memory errors when using the $16e^{-4}$ setting, which prioritize compression for smaller file sizes at the expense of quality. While FCGS offers flexibility in adjusting quality and file size trade-offs, its quality-focused configuration ($1e^{-4}$) lags behind \name in both compression ratios and time costs, and its high compression ratio setting ($16e^{-4}$), which consumes excessive memory, limits its applicability—particularly on resource-constrained devices like mobile platforms. Additionally, each adjustment in FCGS requires a full feedforward pass that takes 30 to 60 seconds, whereas \name completes the process in just 1 to 2 seconds with one step in \foa.

\begin{table}[h]
    \caption{PSNR scores and model sizes for ablations on \adp and \mpq. Each contributes similarly and complements the other to maximize compression performance together.}
    \centering
    \resizebox{\columnwidth}{!}{ 
    \setlength{\tabcolsep}{4pt}
    \begin{tabular}{l|cc|cc|cc}
        \toprule[0.15em]
        PSNR (dB) $\mid$ Size (MiB) & \multicolumn{2}{c|}{Garden} & \multicolumn{2}{c|}{Truck} & \multicolumn{2}{c}{DrJohnson} \\
        \midrule[0.05em]
        \rowcolor{blue!5} Baseline (\gaussian)~\cite{kerbl20233d} & 27.18 & 1379.99 & 24.94 & 601.03 & 28.94 & 805.36 \\ 

        \midrule[0.05em]

        $+$ \adp & 26.51 & 439.73 & 24.27 & 102.89 & 28.21 & 137.87 \\
        $+$ \mpq & \bestcell{26.81} & 216.65 & \bestcell{24.69} & 93.97 & \bestcell{28.81} & 124.43\\ 
        \rowcolor{pink!20} \name(Ours) & 26.29 & \bestcell{89.58} & 24.10 & \bestcell{21.87} & 28.11 & \bestcell{28.98} \\  
        \bottomrule[0.15em]
    \end{tabular}
    }
    \label{tab:ablation}
\end{table}

\textbf{Ablation study.}
We conduct ablation studies to assess the impact of each component. Our results show that \adp and \mpq must work \textbf{synergistically} to achieve the best compression performance, which is impractical when applying either individually in a training-free setup. We analyze the contributions at the module level (i.e., \adp and \mpq) and then delve deeper into each module to examine the impact of various design decisions and strategies. \tref{tab:ablation} compares the compressed models in rendering quality and file size, using one scene from each dataset across four configurations: the original \gaussian~\cite{kerbl20233d} on our hardware setup, \adp individually, \mpq individually, and both combined. While \mpq achieves a slightly higher compression ratio and less quality loss than \adp, further reducing bit-widths in \mpq compromises quality significantly. Combining \adp with \mpq unlocks an additional 2.4 - 4.3$\times$ compression ratio while satisfying the specified quality constraints.

\begin{table}[h]
    \caption{PSNR and model sizes for ablations on attribute-discriminative pruning, which boosts compression ratios while preserving quality over attribute-agnostic pruning.}
    \huge
    \centering
    \resizebox{\columnwidth}{!}{ 
    \begin{tabular}{l|cc|cc|cc}
        \toprule[0.15em]
        
        PSNR (dB) $\mid$ Size (MiB) & \multicolumn{2}{c|}{Garden} & \multicolumn{2}{c|}{Truck} & \multicolumn{2}{c}{DrJohnson} \\
        \midrule[0.05em]
        \rowcolor{blue!5} Baseline (\gaussian)~\cite{kerbl20233d} & 27.18 & 1379.99 & 24.94 & 601.03 & 28.94 & 805.36 \\
        \midrule[0.05em]
        Row-P  & 25.10 & 552.00 & 21.82 & 120.21 & 26.53 & 161.07 \\

        SH-P  & 23.71 & \bestcell{327.46} & 22.90 & 142.62 & 27.40 & 191.10 \\

        \rowcolor{pink!20} \adp & \bestcell{26.51} & 439.73 & \bestcell{24.27} & \bestcell{102.89} & \bestcell{28.21} & \bestcell{137.87} \\ 

        \bottomrule[0.15em]
    \end{tabular}
    }
    \label{tab:ablation_adp}
\end{table}

\textbf{How is being aware of attribute sensitivity significant in pruning?}
\adp prunes both Gaussian primitives and their attributes. \tref{tab:ablation_adp} compares pruning along each dimension (Row-P and SH-P) with \adp in terms of rendering quality (PSNR scores) and file sizes across scenes from three datasets, all with quantization deactivated. \adp clearly outperforms in preserving quality at similar compression ratios, as further supported by curves in \fref{fig:mixed-dim}.

\begin{table}[h]
    \caption{Quality impact of subchannel-wise grouped quantization, halving quality loss at the same compression ratio.}
    \centering
    \huge
    \resizebox{\columnwidth}{!}{
    \begin{tabular}{l|ccc|ccc}
        \toprule[0.15em]

        ~ & \multicolumn{3}{c|}{Garden} & \multicolumn{3}{c}{Bonsai} \\
        ~ & $PSNR^\uparrow$ & $SSIM^\uparrow$ & $LPIPS^\downarrow$ & $PSNR^\uparrow$ & $SSIM^\uparrow$ & $LPIPS^\downarrow$ \\
        \midrule[0.05em]
        \rowcolor{blue!5} Baseline (\gaussian)~\cite{kerbl20233d} & 27.18 & 0.861 & 0.115 & 31.98 & 0.938 & 0.208 \\ 
        
        \midrule[0.05em] 
        
        w/o Group Quant.  & 26.45 & 0.832 & 0.148 & 30.25 & 0.915 & 0.237 \\
        
        \rowcolor{pink!20} w/ Group Quant. & \bestcell{26.81} & \bestcell{0.846} & \bestcell{0.131} & \bestcell{31.19} & \bestcell{0.928} & \bestcell{0.219} \\ 

        \bottomrule[0.15em]
    \end{tabular}
    }
    \label{tab:ablation_mpq}
\end{table}

\textbf{Does subchannel-wise grouped quantization in \mpq improve rendering quality?} \textbf{Yes.} \tref{tab:ablation_mpq} compares the rendering quality of scenes from two datasets, both with and without grouped quantization, while keeping pruning deactivated. Subchannel-wise grouped quantization significantly reduces quantization error, mitigating about half of the quality loss at the same compression ratio. We omit file sizes as this design choice has a negligible impact.

\textbf{How does importance score calculation impact compression performance?}
While we use LightGaussian~\cite{fan2023lightgaussian} for importance scores, \name is independent of score calculation methods. To verify this, we test scores calculated from MesonGS~\cite{xie2025mesongs} and compare rendering quality and compression ratios across scenes from three datasets in \tref{tab:generalizability}. The results show that our system remains effective while meeting compression constraints. The performance gap indicates that more accurate importance estimation can improve both compression ratio and quality.
\begin{table}[h]
    \caption{PSNR and model sizes of \name with importance scores from MesonGS, demonstrating its compatibility.}
    \centering
    \resizebox{\columnwidth}{!}{
    \begin{tabular}{l|cc|cc|cc}
        \toprule[0.15em]
    
        PSNR (dB) $\mid$ Size (MiB) & \multicolumn{2}{c|}{Garden} & \multicolumn{2}{c|}{Truck} & \multicolumn{2}{c}{DrJohnson} \\
        \midrule[0.05em]
        \rowcolor{blue!5} Baseline (\gaussian)~\cite{kerbl20233d} & 27.18 & 1379.99 & 24.94 & 601.03 & 28.94 & 805.36 \\ 

        \midrule[0.05em]

        MesonGS~\cite{xie2025mesongs} & 26.30 & \bestcell{80.40} & 23.99 & \bestcell{21.87} & \bestcell{28.24} & \bestcell{28.98} \\ 
        \rowcolor{pink!20} \name(Ours) & \bestcell{27.17} & 89.80 & \bestcell{24.16} & 21.91 & 28.13 & 29.19 \\  

        \bottomrule[0.15em]
    \end{tabular}
    }
    \label{tab:generalizability}
\end{table}

\section{Conclusion and Future Work}
\label{sec:conclusion}

We propose \name, a flexible, cost-efficient, and training-free method for compressing \gaussian models with high rendering quality, high compression ratios, and low adaption costs to varying compression demands. At its core are attribute-discriminative pruning, ultra-low bit mixed-precision quantization, and a fast online adaptation algorithm that supports dynamically adjusting the compression ratio. \name achieves up to \fixme{96.4\%} compression with $<$1 dB PSNR drop in just seconds --- two orders of magnitude faster than prior retraining- and refinement-based approaches. \name is deployable on resource-constrained mobile devices, which is impractical for all other compression methods. For future work, we plan to explore further pushing the compression ratio by investigating ternary or even binary attributes as well as achieving even faster adaption speed via advanced online search algorithms.

\section*{Acknowledgments}

We sincerely appreciate the insightful feedback from the anonymous reviewers. We also thank Professor Sarita Adve of the University of Illinois Urbana-Champaign for her valuable input and continued support throughout this research. This research was supported by the National Science Foundation (NSF) under Grant No. 2441601. The work utilized the Delta and DeltaAI system at the National Center for Supercomputing Applications (NCSA) through allocation CIS240055 from the Advanced Cyberinfrastructure Coordination Ecosystem: Services \& Support (ACCESS) program, which is supported by National Science Foundation grants \#2138259, \#2138286, \#2138307, \#2137603, and \#2138296. The Delta advanced computing resource is a collaborative effort between the University of Illinois Urbana-Champaign and NCSA, supported by the NSF (award OAC 2005572) and the State of Illinois. This work also utilized the Illinois Campus Cluster and NCSA NFI Hydro cluster, both supported by the University of Illinois Urbana-Champaign and the University of Illinois System. UIUC SSAIL Lab is supported by research funding and gift from Google, IBM, and AMD.


\bibliographystyle{ACM-Reference-Format}
\bibliography{reference}



\end{document}